\documentclass{article}
\usepackage{graphicx} % Required for inserting images
\usepackage[a4paper, total={6in, 8in}]{geometry}
\usepackage{authblk}
\usepackage{footmisc}
\usepackage{booktabs}
\usepackage{tcolorbox}
\tcbuselibrary{skins,breakable}
\usepackage{hyperref}
\usepackage{multirow}
\usepackage[automake]{glossaries}
\usepackage{subfig}
\usepackage{caption}
\usepackage{makecell}
\usepackage{soul}

\usepackage{floatrow}
\newfloatcommand{capbtabbox}{table}[][\FBwidth]

\usepackage{framed}
\usepackage{xcolor}

\usepackage{amssymb}
\makeatletter
\newcommand{\Rmnum}[1]{\expandafter\@slowromancap\romannumeral #1@}
\makeatother
\usepackage{tabularx}

\usepackage{float}
\newfloat{figtab}{htb}{fgtb}
\makeatletter
  \newcommand\figcaption{\def\@captype{figure}\caption}
  \newcommand\tabcaption{\def\@captype{table}\caption}
\makeatother

\usepackage{url}

\makeglossaries
\glsaddall
\newacronym{GPT-3}{GPT-3}{Generative Pre-trained Transformer 3}
\newacronym{GPT}{GPT}{Generative Pre-trained Transformer}
\newacronym{LLMs}{LLMs}{Large Language Models}
\newacronym{STPA}{STPA}{Systems Theoretic Process Analysis}
\newacronym{STAMP}{STAMP}{System-Theoretic Accident Model and Processes}
\newacronym{AEB}{AEB}{Automatic Emergency Brake}
\newacronym{T-PTLMs}{T-PTLMs}{Transformer-based pretrained language models}
\newacronym{NLP}{NLP}{Natural Language Processing}
\newacronym{GPT-1}{GPT-1}{Generative Pre-Training Transformer-1}
\newacronym{BERT}{BERT}{Bidirectional Encoder Representations from Transformers}
\newacronym{XLNet}{XLNet}{Transformer-XL Network}
\newacronym{RoBERTa}{RoBERTa}{Robustly Optimized BERT Pretraining Approach}
\newacronym{ELECTRA}{ELECTRA}{Efficiently Learning an Encoder that Classifies Token Replacements Accurately}
\newacronym{T5}{T5}{Text-to-Text Transfer Transformer}
\newacronym{ALBERT}{ALBERT}{A Lite BERT}
\newacronym{BART}{BART}{Bidirectional and Auto-Regressive Transformers}
\newacronym{PEGASUS}{PEGASUS}{Pre-training with Extracted Gap-sentences for Abstractive Summarization}
\newacronym{UCAs}{UCAs}{Unsafe Control Actions}
\newacronym{HMI}{HMI}{Human Machine Interface}
\newacronym{DSM}{DSM}{Demand Side Management}

\newcount\Comments  % 0 suppresses notes to selves in text
\usepackage{color}
\definecolor{darkgreen}{rgb}{0,0.5,0}
\definecolor{purple}{rgb}{1,0,1}
\newcommand{\kibitz}[2]{\ifnum\Comments=1\textcolor{#1}{#2}\fi}
\newtheorem{remark}{Remark}

\title{Safety Analysis in the Era of Large Language Models: \\ A Case Study of STPA using ChatGPT }

\author{Yi~Qi$^1$, Xingyu~Zhao$^{1,2}$, Siddartha Khastgir$^2$, Xiaowei~Huang$^1$
\thanks{Emails:\{yi.qi,xiaowei.huang\}@liverpool.ac.uk, \{xingyu.zhao,s.khastgir.1\}@warwick.ac.uk}
}
\affil{$^1$Computer Science Department, University of Liverpool, Liverpool L69 3DR, U.K.
\\
$^2$WMG, University of Warwick, Coventry CV4 7AL, U.K.
}

\begin{document}

\date{}
\maketitle

\begin{abstract}
Can safety analysis make use of Large Language Models (LLMs)? A case study explores the application of Systems Theoretic Process Analysis (STPA) to Automatic Emergency Brake (AEB) and Electricity Demand Side Management (DSM) systems, using ChatGPT. We investigate how collaboration schemes, input semantic complexity, and prompt engineering influence STPA results. Comparative results show that using ChatGPT without human intervention may be inadequate due to reliability related issues, but with careful design, it may outperform human experts. No statistically significant differences are found when varying the input semantic complexity or using domain-agnostic prompt guidelines. While STPA-specific prompt engineering shows statistical significance of producing more pertinent results, ChatGPT tends to produce more conservative and less comprehensive outcomes. We also highlight future challenges, including concerns about LLM trustworthiness and the necessity for standardisation and regulation in this domain. All our experimental data are publicly accessible.

%To answer, a case study of Systems Theoretic Process Analysis (STPA) on Automatic Emergency Brake (AEB) systems and Electricity Demand Side Management (DSM) systems using ChatGPT is conducted. Spanning from a broad overview to detailed granularity, we abstract three levels of interactions that experts may have with ChatGPT, namely workflow, semantics and syntax levels. Consequently, we investigate research questions regarding how workflow-level collaboration schemes, the semantic complexity of input questions, and syntactic prompt guidelines influence the STPA results. Comparative results showed that using ChatGPT without human experts’ intervention can be inadequate due to its accuracy issues. However, using ChatGPT in STPA with extra care can outperform human safety experts alone. Additionally, no statistically significant differences were found to indicate that input semantics complexity levels and the use of a generic prompt guideline impact ChatGPT's outputs. This paper also identifies challenges for future research in this direction, such as trustworthiness concerns of LLMs and the need for standardisation and regulation.
\end{abstract}

\textbf{Keywords}:  STPA, safety-critical systems, ChatGPT, large language models, safe AI, automatic emergency brake systems, human machine interaction, hazards identification, safety assurance.

\section{Introduction}

\gls{LLMs} \cite{keskar2019ctrl}, including \gls{GPT} \cite{NEURIPS2020_1457c0d6} and \gls{BERT} \cite{devlin-etal-2019-bert}, have achieved state-of-the-art performance on a wide range of \gls{NLP} tasks. \gls{LLMs} are gaining popularity and receiving increasing attention for their significant applications in knowledge reasoning \cite{dasgupta2022language,yasunaga2021qa,https://doi.org/10.48550/arxiv.2205.10625}. ChatGPT is one of the \gls{LLMs} applications, and probably \textit{the} application, in the limelight. ChatGPT was used for collating literature and writing professional papers in fields like law \cite{choi2023chatgpt}, and medical education \cite{kung2023performance,gilson2023does}. 
OpenAI announced GPT-4 in March 2023 that can pass some of the bar exams to AP Biology \cite{openai2023gpt4}.
These successful stories demonstrate that people have already gained experience in using \gls{LLMs}, for their performance in handling complex content due to their massive training datasets and model capacity to process and learn from data, enabling their potential for complex tasks that require domain expert knowledge \cite{min2021recent}. Given this, as researchers in the field of safety-critical systems, we pose a question: \textit{Can safety analysis make use of \gls{LLMs}?}
% \yiqi{One sentence: What functions or analyses does STPA perform?}

The \gls{STAMP} is a safety analysis framework based on system theory, with \gls{STPA} being one of the most prevalent STAMP-based tools \cite{leveson2011engineering}. 
It considers the entire system, including interactions among components and human operators, to identify potential hazards \cite{leveson2018stpa}.
%It includes an explicit system modelling step, which is useful for structuring the input to the LLMs.\xingyu{unclear }
Despite being a promising approach, \gls{STPA} (or, more generally, all safety analysis methods) is subject to limitations, including high complexity (especially when applied on modern systems with heterogeneous components ranging from hardware, software, UI/UX, human operators, to AI/ML models) and subjectivity (i.e., subjective opinions of experts conducting STPA can lead to inconsistent, imprecise or biased results) \cite{DAKWAT2018130,harkleroad2013review,abdulkhaleq2013experiences,10.1007/978-3-319-45480-1_16}. LLMs may have the potential to cope with them, yet their effectiveness is uncertain. This case study is designed to investigate their actual potential and explore ways to realise it effectively.

\paragraph{Motivation} 
More specifically, we are motivated by two primary objectives: to examine how LLMs like ChatGPT can assist in navigating the complexities of intricate systems, and to investigate whether LLMs can bring a broader knowledge base (thereby increasing objectivity) into safety analysis, complementing human expertise.
\begin{figure}[htp]
\centering
\includegraphics[width=1\linewidth]{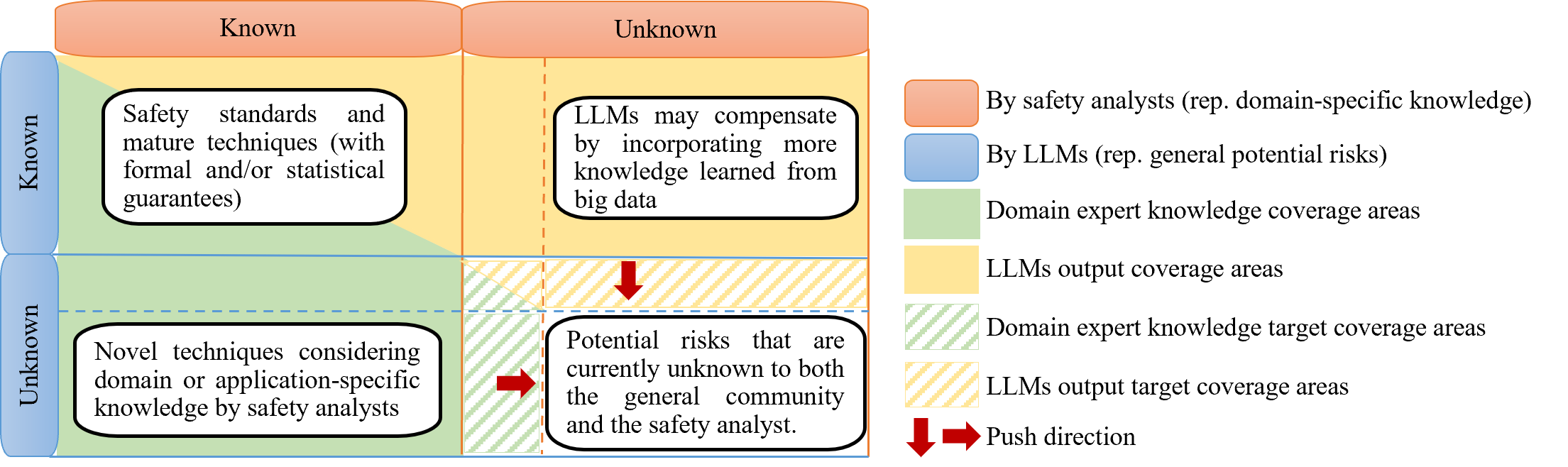}
\caption{Four-quadrant classification of risks with ways of mitigations}
\label{fig4}
\end{figure}

Fig.~\ref{fig4} shows the four-quadrant classification of safety risks depends on if the risk is known by the safety analysts (representing domains-specific knowledge) or the LLMs (representing general knowledge from big data). Consequently, we derive the four types of risk as:
\begin{itemize}
    \item Known-known (top-left): Risks associated with the most common hazards that are generally known and also cannot be overlooked by any safety analysts when doing safety analysis. To mitigate them, the safety-critical system community has developed safety standards 
    %(e.g., \cite{iec61508_2010}) 
    and mature verification and validation techniques (yielding formal and/or statistical guarantees).
    \item Known-unknown (top-right): Risks well-known in general, e.g., those documented in standards or discussed in academic literature, can sometimes be inadvertently ignored in specific system safety analyses. LLMs have the potential to mitigate this oversight. They can do so by integrating extensive knowledge derived from big data, thus ensuring a more comprehensive consideration of known risks in safety analysis.
    \item Unknown-known (bottom-left): Risks that are too novel (due to, e.g., emerging techniques like AI/ML) to be known by the general community yet, and domain-specific (or even application-specific) that can only be studied in a case by case manner by the safety analysts. Their mitigation requires novel techniques considering domain/application-specific knowledge \cite{bloomfield_disruptive_2019}.
    \item Unknown-unknown (bottom-right): Potential risks that are currently unknown to both the general community (represented by big data) and the domain-specific human safety experts. 
\end{itemize}
The ultimate goal of the safety community is to reduce the area of ``Unknown-unknown'' (bottom-right), we can see in the big picture in Fig.~\ref{fig4} that how the LLMs and human safety experts can jointly push the boundaries (horizontal and vertical bars respectively) to achieve this goal.

\paragraph{Approach} We conduct a case study of \gls{STPA} on \gls{AEB} systems and \gls{DSM} systems using ChatGPT. 
To enable comparative studies, we reuse the analysis result completed by an human experts, published in \cite{10.1115/1.4051940} and \cite{karatzassystem}, as baselines. Ranging from coarse to fine-grained abstraction levels, we examine the interactions between the human analysts and ChatGPT at: workflow, semantics, and syntax levels. At the workflow level, we assess factors such as the frequency, direction, and timing of interactions between the human analysts and ChatGPT. For individual prompts at the semantics level, we evaluate the influence of input complexity on ChatGPT's output. Finally, at the syntax level, we study if the adoption of domain-agnostic and STPA-specific prompt engineering helps or not.

\paragraph{Key Findings and Contributions}According to both qualitative comparison and statistical analysis, some key findings emerge: 
\begin{itemize}
    \item Applying ChatGPT without human experts' intervention is inadequate due to the problematic reliability of LLMs in general.
    \item More co-analysis activities and human-ChatGPT interactions in STPA may outperform human experts alone.
    \item Neither input complexity nor the use of domain-agnostic prompts had a statistically significant impact on ChatGPT's outputs.
    \item While STPA-specific prompt engineering shows statistical significance of producing more pertinent results, ChatGPT tends to produce more conservative and less comprehensive outcomes.
    %due to the limitations imposed by the more constrained prompts.
\end{itemize}

% While the ultimate goal of the safety community is to reduce 
% the area of ``Unknown-Unknown'', we can see clearly in the big picture in Fig.~\ref{fig4} that how the LLMs and human safety experts can jointly push the boundary (horizontal and vertical bars) to achieve this goal (bottom-right). The utilisation of \gls{LLMs} in safety analysis can enable human experts to integrate knowledge learned from big data. This integration helps reduce subjectivity that might be introduced by individual experts. Moreover, modern systems have become increasingly complex, encompassing heterogeneous components ranging from hardware, software, UI/UX, to AI/ML models. Therefore, safety experts need a more diverse knowledge base than ever before to conduct a comprehensive safety analysis. We believe that assistance in safety analysis from automated tools like LLMs, which can identify potential safety issues based on historical and big data, may be very useful when analysing such complex modern systems.

We summarise our contributions as follows: A first attempt on applying \gls{LLMs} in safety analysis for safety-critical systems is made. We develop and compare various ways to interact with ChatGPT in STPA, then make remarks and draw conclusions regarding what determines effectiveness and what does not. All our findings are based on evidence generated from comparative studies and statistical analysis, with threats to validity discussed. Opportunities of safety analysis using \gls{LLMs} are discussed, while key challenges are identified as research questions, paving the way for future research in this direction. 

% \yiqi{Re-write motivation: refer to sec.5.1}

\section{Background}

\subsection{Large Language Models}

LLMs are advanced machine learning  models that are designed to understand natural language and generate human-like responses to queries and prompts. LLMs are among the most impressive achievements in AI research and development. Almost all \gls{NLP} tasks (e.g., language translation, chatbots and virtual assistants) have witnessed tremendous success with \gls{T-PTLMs}, relying on Transformer \cite{NIPS2017_3f5ee243}, self-supervised learning \cite{jaiswal2020survey,liu2021self} and transfer learning \cite{houlsby2019parameter,ruder2019transfer} to process and understand the nuances of human language, including grammar, syntax, and context.

The success of \gls{NLP} tasks owes a great deal to T-PTLMs such as GPT-1 \cite{radford2018improving}, BERT \cite{devlin-etal-2019-bert}, XLNet \cite{yang2019xlnet}, RoBERTa \cite{liu2019roberta}, ELECTRA \cite{clark2020electra}, T5 \cite{raffel2020exploring}, ALBERT \cite{lanalbert}, BART \cite{lewis-etal-2020-bart}, and PEGASUS \cite{pmlr-v119-zhang20ae}. These models have the capability to learn general language representations from large volumes of unlabelled text data through self-supervised learning, and subsequently transfer this knowledge to specific tasks, which has been a major factor contributing to their success in \gls{NLP} \cite{kalyan2021ammus}. Kaplan et al. \cite{kaplan2020scaling} demonstrated that increasing the size of T-PTLMs can lead to improved performance \cite{kalyan2021ammus}. This finding has spurred the development of \gls{LLMs} such as GPT-3 \cite{NEURIPS2020_1457c0d6}, PANGU \cite{https://doi.org/10.48550/arxiv.2104.12369}, GShard \cite{lepikhin2020gshard} and Switch-Transformers \cite{fedus2021switch}. One model that has gained significant attention is GPT-3, and its improved version GPT-3.5 that forms the technical core of ChatGPT. At the time of writing this paper, ChatGPT has
integrated with GPT-4. 
There are emerging surveys on LLMs from diverse perspectives, including performance \cite{zhao2023survey}, verification and validation \cite{huang2023survey}, and explainability \cite{zhao2023explainability}.

\subsection{Systems Theoretic Process Analysis}
\label{sec_stpa}

\gls{STPA} is a widely used safety analysis method that is based on the \gls{STAMP} model. The main focus of \gls{STPA} is on the overall control loop and process analysis of a system, with an emphasis on identifying unsafe control actions and causal factors within the control structure. This method has found applications in various domains, including railway safety assurances \cite{yang2019automated}, cyber safety and security \cite{kaneko2018threat}, collaborative robots \cite{adriaensen2021stpa}, and autonomous vehicles \cite{khastgir2021systems,chen2020identifying}. Readers are refereed to \cite{harkleroad2013review,qi2023stpa}.

The workflow of STPA, as outlined by N. Leveson \cite{leveson2018stpa}, encompasses five critical steps:
% The \gls{STPA} workflow includes the following steps \cite{leveson2018stpa}:  
1) Define Accidents/Hazards. 2) Model Control Structure. 3) Identify Unsafe Control Actions. 4) Identify Causal Scenarios. 5) Derive Safety Requirements.
Specifically, the initial step involves defining potential accidents or hazards at a high level, typically encompassing significant negative outcomes for personnel or equipment that can be anticipated \cite{ishimatsu2010modeling}. The subsequent step involves creating a system control structure that segregates the control loop from its individual components, such as the starting point, intermediate processes (like mechanical and software structures), and completion components. 
Following this, 
% standard \gls{UCAs} form with four categories, 
% including Causes Hazards Not Providing, Causes Hazard Providing, Incorrect Timing or Order, and Stopped Too Soon or Applied Too Long, can be employed to identify possible \gls{UCAs}. 
a standard \gls{UCAs} consist of the following parts:
\begin{equation}
\label{eq_uca_parts}
    Source + Type + Control\; Action + Context + Link\; to\; Hazards
\end{equation}
This structure of UCAs
% as delineated by N. Leveson and J. Thomas in \cite{leveson2018stpa}, 
embodies a comprehensive framework that necessitates meticulous consideration of each constituent element\footnote{The first element comprises the controller responsible for delivering the control action. Following this, the second element categorises the nature of the unsafe control action, which may manifest as either provided, not provided, too early or too late, stopped too soon or applied too long. The third element pertains to the specific control action or directive, as defined within the control structure. The fourth element addresses the previously mentioned context, while the final element establishes a connection to hazards or sub-hazards. Although UCAs are typically documented in the order outlined above, alternative sequencing might be more coherent or intuitive in certain instances. The sequence itself is not of paramount importance; rather, the inclusion of all five components is crucial.}.
After analysing the \gls{UCAs} to identify potential hazards or risks, the \gls{STPA} proceeds to determine potential causes and scenarios. The final step involves generating safety requirements for each identified risk or hazard.

Expert knowledge is essential in order to conduct all five steps described above. The first step largely depends on the safety-critical application and its environments, while the second step rely on the system's internal structure. The third step is highly dependent on expertise of the safety analyst, as it requires an in-depth understanding of the system or process under analysis, as well as prior knowledge of relevant safety regulations and standards. The fourth and fifth steps also require the expert knowledge, as the expert team is able to identify and evaluate potential failure modes and develop reasonable safety requirements.
Despite its popularity, conducting STPA is widely acknowledged to be challenging. 
The complexity arises from the increasingly complex systems being analysed (including recent learning-enabled systems \cite{qi2023stpa}), while subjectivity is due to the limited knowledge and expert judgement of individuals \cite{DAKWAT2018130}. These motivate us to investigate if and how LLMs can address the two challenges.

\section{Methodology}

%According to how an expert team may use ChatGPT for STPA, 
Our case study comprises a series of experiments, conducting both qualitative and quantitative analyses on the output generated by ChatGPT in STPA. Two independent human expert teams were engaged to participate in and independently validate our experimental results.

\subsection{Research Questions}
We first define three levels of abstraction, ranging from coarse to fine-grained, that represents how human experts may interact with ChatGPT:
\begin{itemize}
    \item \textbf{Workflow Level}: At this most coarse abstraction level, we examine the varying frequencies, directions, and timing of expert interactions with ChatGPT throughout the STPA process. 
    \item \textbf{Semantics Level}:
    Given a workflow, the system under analysis can be modelled with varying amount of semantic information, i.e., level of system details in the control loop structures. We consider such differences in input complexity and study the impact on outputs.
    \item \textbf{Syntax Level}: 
    While preserving the main semantic information, the order of words, sentence structures and requirements on output formats may vary in individual prompts, with or without prompt engineering. We consider such syntactic differences at this level. 
\end{itemize}
To make remarks and draw conclusions on the optimal ways of using ChatGPT for STPA, we frame the following research questions (RQs), corresponding to the three %abstraction 
levels of collaboration, respectively:
\begin{itemize}
    \item RQ1 (Collaboration Scheme): How do various collaboration schemes of integrating ChatGPT into STPA affect the effectiveness and usability of STPA?
    \item RQ2 (Control Loop Complexity): To what extent do variations in semantic complexity of individual input questions to ChatGPT affect the comprehensiveness and pertinence of STPA results?
    %affect the proportion of correct UCAs and quantity of correct UCAs in STPA?
    % To what extent do variations in semantic complexity of the input questions to ChatGPT affect the accuracy and comprehensiveness\xingyu{better match to those two statstics we studied..} of its output in STPA?
    \item RQ3 (Prompt Engieering): Does the utilisation of domain-agnostic and STPA-specific prompt engineering affect the comprehensiveness and pertinence of STPA results?
    % Does the utilisation of syntactic-level prompt engineering significantly impact\xingyu{in terms of what?} the outcomes of STPA?
\end{itemize}

\subsection{Systems Under Study}

In our case studies, we have selected two safety-critical systems and reuse their \gls{STPA} results from \cite{10.1115/1.4051940} and \cite{karatzassystem} as our baselines (which were obtained from human safety experts).

\subsubsection{Baseline \Rmnum{1}: Automatic Emergency Brake Systems}

\paragraph{AEB Systems}
If a collision is about to occur and the driver takes no action or the action is not fast enough, the \gls{AEB} system will automatically initiate braking. \gls{AEB} is able to detect potential collisions and activate the braking system to slow down the vehicle to avoid the collision or %at least 
reduce its impact.
A typical \gls{AEB} system consists of many components, including signal acquisition, calculation, algorithm, and fusion processes, as well as interfaces with electrical and mechanical parts, sensor systems, and more. AEB is responsible for detecting hazards, making decisions, and applying brakes to avoid or mitigate collisions.

%We compile and present some of the STPA findings from \cite{10.1115/1.4051940}%\footnote{There are four parts consisting of a hazardous control action of each UCA \cite{qi2023stpa}, i.e., \textit{Source Controller}, \textit{Type}; and \textit{(T4) The control action lasts too long or is stopped too soon},  \textit{Control Action}, and \textit{Context}. The four \textit{Type}s in UCAs are: \textit{(T1) Not providing the control action leads to a hazard}; \textit{(T2) Providing the control action leads to a hazard}; \textit{(T3) Providing a potentially safe control action but too early, too late, or in the wrong order}. }

\begin{table}[h]
\caption{Accidents, hazards and example UCAs for one control action from \cite{10.1115/1.4051940}}
\label{tab: accidents_aeb}
\scalebox{0.88}{
\begin{tabular}{l|l}
\toprule
Accidents/Losses                    & Hazards (\textit{H}) \\ 
\midrule
Loss of property    & No brake, faulty brake (insufficient or excessive), or misuse of the brake provided. (\textit{H1})\\ 
Human damage    & Sensor signal/monitoring inadequate, inaccurate, delay or missing. (\textit{H1})\\    
\bottomrule
\toprule
% \hline\hline
Control Action                    & Unsafe Control Actions (UCAs) \\ 
\midrule
\multirow{4}{*}{\makecell*[l]{Implementation of \\ brake commands}} & No brake command provides a cause of a hazard. (\textit{H1, H2})\\ 
    & The brake command is inadequately provided under no collision risk. (\textit{H1, H2})\\    
    & Brake command occurring too late causes insufficient brake and crash risk. (\textit{H1, H2})\\  
    & Braking too long or too short causes a hazard. (\textit{H1})\\
\bottomrule
\end{tabular}
}
\end{table}

% \begin{table}[h]
% \caption{Example UCAs for one control action from \cite{10.1115/1.4051940}}
% \label{tab: uca_aeb}
% \scalebox{0.88}{
% \begin{tabular}{l|l}
% \toprule
% Control Action                    & Unsafe Control Actions (UCAs) \\ 
% \midrule
% \multirow{4}{*}{\makecell*[l]{Implementation of \\ brake commands}} & No brake command provides a cause of a hazard. (\textit{H1, H2})\\ 
%     & The brake command is inadequately provided under no collision risk. (\textit{H1, H2})\\    
%     & Brake command occurring too late causes insufficient brake and crash risk. (\textit{H1, H2})\\  
%     & Braking too long or too short causes a hazard. (\textit{H1})\\
% \bottomrule
% \end{tabular}
% }
% \end{table}

\paragraph{STPA Results by Human Experts}
% \xingyu{Please metion all 5 steps of the STPA, starting from table 1, and metnion control loop structure, then UCAs table 2, and then say step 4, 5 omiited ...}

As per section \ref{sec_stpa}, STPA typically encompasses five steps. The first step involves defining accidents/hazards. As illustrated in Table \ref{tab: accidents_aeb}, we have extracted the accidents and hazards associated with AEB systems as given by experts in paper \cite{10.1115/1.4051940}. The second step entails identifying the control structure (cf. later Figure \ref{fig:baseline_loop_com}). The third step seeks to identify UCAs within the control loop, as detailed in Table \ref{tab: accidents_aeb}.
% The work \cite{10.1115/1.4051940} presents a list of \gls{UCAs} for the control action ``implementation of brake commands'', as shown in Table \ref{tab: uca_aeb}. 
% %It includes a correspondence between each UCA and its corresponding UCA type. 
% The next step 4 in \gls{STPA} is to analyse and identify possible casual scenarios for the four \gls{UCAs}. For example, for ``Braking too long or too short causes a hazard'', AEB control model processing delays result from the braking being too long is a possible scenarios and causes\xingyu{A bit hard to understand this example?}. Finally, safety requirements can be derived for the causes identified. Note, the last two STPA steps are omitted in paper \cite{10.1115/1.4051940}.
%as the focus is on comparing safety analysis methods, 
We have selected ``implementation of brake commands" as an example. Pertaining to this control action, there are four UCAs. In the fourth step of STPA, the objective is to ascertain the causal scenarios for each UCA. Subsequently, the final step involves proposing safety requirements in response to these causal scenarios.
Note, the last two STPA steps are omitted in paper \cite{10.1115/1.4051940}.
Similarly, we also omit the last 2 steps in our later experiments and mainly focus on finding UCAs.

\subsubsection{Baseline \Rmnum{2}: Electricity Demand Side Management Systems}

\paragraph{Electricity DSM Systems}
Electricity \gls{DSM} involves adjusting the power consumption patterns of end-users based on fluctuations in electricity prices or incentive payments. These adjustments are made to mitigate system failures and reduce overall electricity consumption when system reliability is at risk. The primary components of this system include Admission Control, Load Balancing, the Demand/Response Manager (DRM), and the Load Forecaster.

\paragraph{STPA Results by Expert Teams}

We show part of the STPA results from \cite{karatzassystem} when it is applied to the DSM system. The accidents and hazards are shown in Table \ref{tab: accidents}. 
% \xingyu{mention the step 2 here using one sentence.. to make sure readers without knowledge of STPA can roughly follow the process.}
Next, expert teams abstract the control loop structure of the systems (cf. later Figure \ref{fig:baseline_loop_com}).
Then, UCAs are identified and shown in Table \ref{tab: accidents}, where we use the UCAs for the control action ``excess capacity demand'' as an example. 
%This control action will serve as our example for comparison. 
Again, we omit the last two STPA steps of identifying casual scenarios and safety requirements.

\begin{table}[h]
\caption{Accidents, hazards and example UCAs for one control action from \cite{karatzassystem}}
\label{tab: accidents}
\scalebox{0.855}{
\begin{tabular}{l|l}
\toprule
Accidents/Losses                    & Hazards (\textit{H}) \\ 
\midrule
Power shortages  & Smartgrid has an inability to meet unexpected demands (\textit{H1})\\ 
Loss of customers    & Smartgrid is unable to satisfy local energy demands (\textit{H2})\\    
Loss of grid equipment    & Smartgrid has an inability to keep customers comfortable per their preferences (\textit{H3})\\

\bottomrule
\toprule
Control Action                    & Unsafe Control Actions (UCAs) \\ 
\midrule
\multirow{6}{*}{Excess capacity demand}  & DRM does not provide capacity limits when these have been modified. (\textit{H2, H3})\\ 
    & \makecell[l]{DRM demands more excessive capacity than the actual required capacity for \\appliances to operate in the defined time horizon ahead.} (\textit{H1})\\    
    & \makecell[l]{DRM demands less excessive capacity than the actual required capacity for \\appliances to operate in the defined time horizon ahead.} (\textit{H2, H3})\\
    & \makecell[l]{DRM demands excessive capacity while the appliances can operate sufficiently\\ in the defined time horizon ahead.} (\textit{H1})\\
    & DRM demands excessive capacity too late (\textgreater TBD) after request. (\textit{H2, H3})\\  
    & DRM stops demanding for excessive capacity while overload still remains. (\textit{H2, H3})\\
\bottomrule

\end{tabular}
}
\end{table}

% \begin{table}[h!]
% \caption{Example UCAs for one control action from \cite{karatzassystem}}
% \label{tab: uca_case2}
% \scalebox{0.86}{
% \begin{tabular}{l|l}
% \toprule
% Control Action                    & Unsafe Control Actions (UCAs) \\ 
% \midrule
% \multirow{6}{*}{Excess capacity demand}  & DRM does not provide capacity limits when these have been modified. (\textit{H2, H3})\\ 
%     & \makecell[l]{DRM demands more excessive capacity than the actual required capacity for \\appliances to operate in the defined time horizon ahead.} (\textit{H1})\\    
%     & \makecell[l]{DRM demands less excessive capacity than the actual required capacity for \\appliances to operate in the defined time horizon ahead.} (\textit{H2, H3})\\
%     & \makecell[l]{DRM demands excessive capacity while the appliances can operate sufficiently\\ in the defined time horizon ahead.} (\textit{H1})\\
%     & DRM demands excessive capacity too late (\textgreater TBD) after request. (\textit{H2, H3})\\  
%     & DRM stops demanding for excessive capacity while overload still remains. (\textit{H2, H3})\\
% \bottomrule
% \end{tabular}
% }
% \end{table}

\subsection{RQ1 Experiment and Evaluation Design}

\begin{figure}[h!]
\centering
\includegraphics[width=\linewidth]{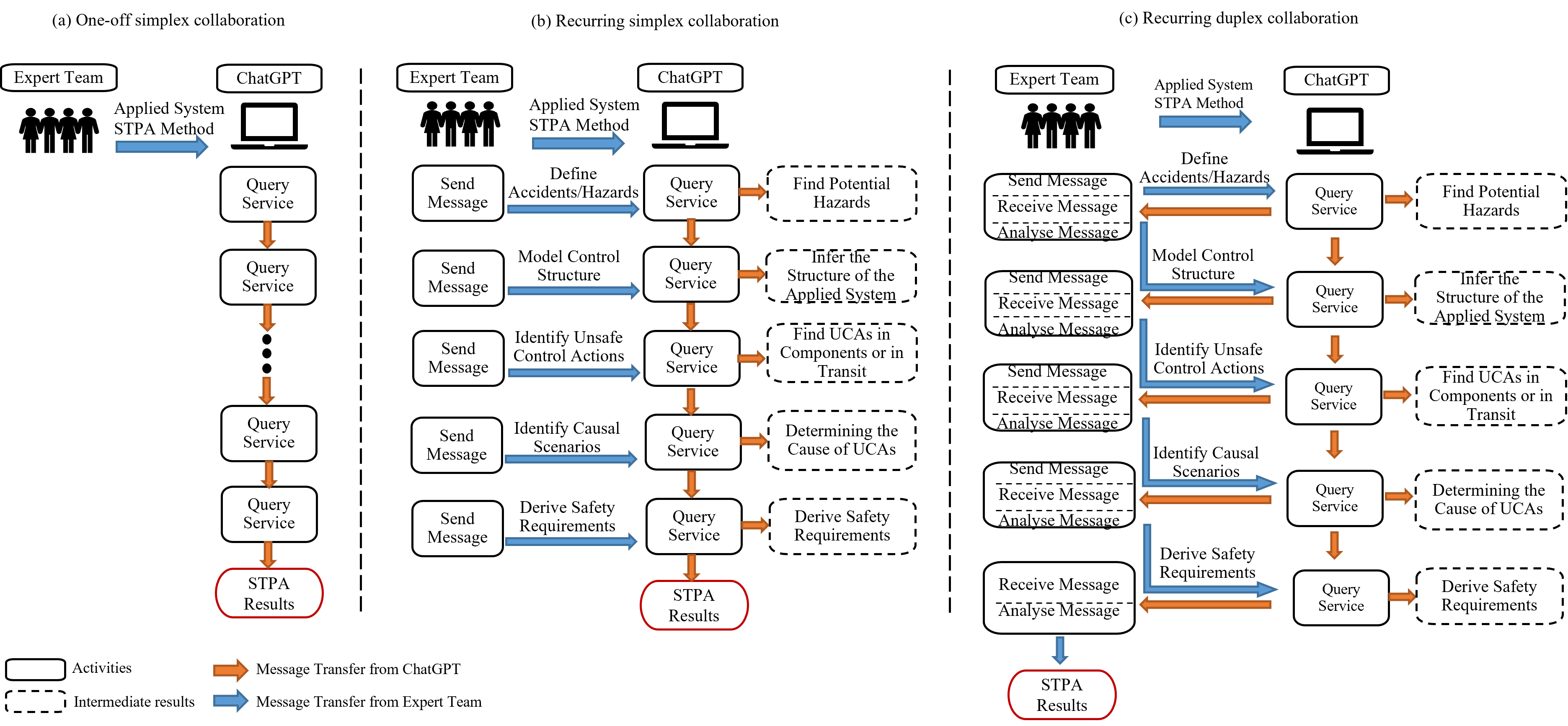}
\caption{Three ways of incorporating ChatGPT in the workflow of how human safety experts perform STPA: (a) One-off simplex collaboration (b) Recurring simplex collaboration (c) Recurring duplex collaboration.}
\label{fig2}
\end{figure}

\paragraph{Experiment Design}
We consider three collaboration schemes of incorporating ChatGPT into the STPA workflow in this work. Specifically, we develop one-off simplex collaboration (Fig.~\ref{fig2}(a)), recurring simplex collaboration (Fig.~\ref{fig2}(b)) and recurring duplex collaboration (Fig.~\ref{fig2}(c)), depending on the communication \textit{frequency} and \textit{direction} between human safety experts and ChatGPT. 

The first case shown in Fig.~\ref{fig2}(a) involves the expert team providing ChatGPT with brief instructions containing the keywords ``AEB System'' and ``STPA Method'', 
%(i.e., the applied system and safety analysis method), 
without sharing the specific details of how to perform \gls{STPA}. Then, STPA is conducted automatically and atomically, according to the steps obtained directly from ChatGPT. This case is referred as ``One-off simplex collaboration'', in the sense that the only collaboration is at the very beginning, and the communication diction is single-directional from human to ChatGPT. Intuitively, this case represents the scenario in which the human experts fully trust ChatGPT and only need to specify their high-level inputs (i.e., conducting STPA on the AEB systems) to get the final STPA results.

The ``Recurring simplex collaboration'' is illustrated in Fig.~\ref{fig2}(b)---information is communicated multiple times in the direction from human experts to ChatGPT. In this case, the expert team begins with providing ``AEB System" and ``STPA Method" to ChatGPT, and then proceeds to perform each of the five steps of STPA. At each step, the experts initiate the analysis by sending clear instructions containing keywords for that step, but without questioning and analysing the answers from the last step. Finally, ChatGPT outputs the overall result of the STPA analysis.

As the name ``Recurring duplex collaboration'' suggests, the third way of integrating ChatGPT in STPA involves multiple bi-directional interactions. As shown in Fig.~\ref{fig2}(c), the human safety experts sending ``AEB System'' and ``STPA Method" to ChatGPT.
However, unlike ``Recurring simplex collaboration'' in Fig.~\ref{fig2}(b), they collect each output of ChatGPT, analyse the information by incorporating domain knowledge, and then query ChatGPT in turn based on the human processed question. The expert team is responsible for ``Sending Messages", ``Reserving Messages", and ``Analysing Messages" in each step of STPA.

\paragraph{Evaluation Design} In addition to noting remarks based on observations during the case studies, we will also reuse the comparison methodology from \cite{10.1115/1.4051940} (designed for a range of safety analysis methods) with an adapted set of attributes to suit our specific goal on comparing STPA results. 
Table \ref{tab3} lists those attributes for comparison, based on which we assess and compare the three schemes of incorporating ChatGPT into STPA process. In addition, the three sets of UCAs obtained through the three collaboration schemes will be also be compared qualitatively using Venn diagrams. 
%Safety experts group will be invited to conduct a qualitative review and summarisation of the results generated by ChatGPT. We will provide ``Remarks'' as a summary of our evaluation process, while also presenting comparison results and Venn diagrams as intuitive visualisations of the comparisons.

\begin{table}[]
\caption{Comparison attribute list (adapted from \cite{10.1115/1.4051940})}
% \yiqi{attribute refer to this table}
\label{tab3}
\scalebox{0.85}{
\begin{tabular}{lll}
\toprule
Aspects                     & Attributes                        & Descriptions \\ 
\midrule
Analysis results comparison & Identify hazards                  & Comprehensiveness of identified hazards  \\ 
Analysis results comparison                           & Identify causes              & Comprehensiveness of causes of identified hazards           \\ 
Analysis process comparison    & Skill required                &  Relative technical experience required as expert teams             \\
Analysis process comparison                            & Time cost &  The time cost required for safety analysis with the case             \\ 
Analysis process comparison   & Complexity/difficulty  & Relative complexity/difficulty of the case structure\\
\bottomrule
\end{tabular}}
\end{table}

\subsection{RQ2 Experiment and Evaluation Design}

\paragraph{Experiment Design}

Within the third collaboration scheme ``Recurring duplex collaboration''\footnote{Because, as the later answer to RQ1 shows, this is the most effective collaboration scheme of using ChatGPT in STPA in our case studies.}, we select STPA steps 2 and 3 to study RQ2, i.e., to examine how various complexity levels of control loop structures derived from step 2 may impact the identified UCAs in step 3.

\begin{figure}[h]
  \centering
   \subfloat[][Low]
  {\includegraphics[width=0.33\textwidth]{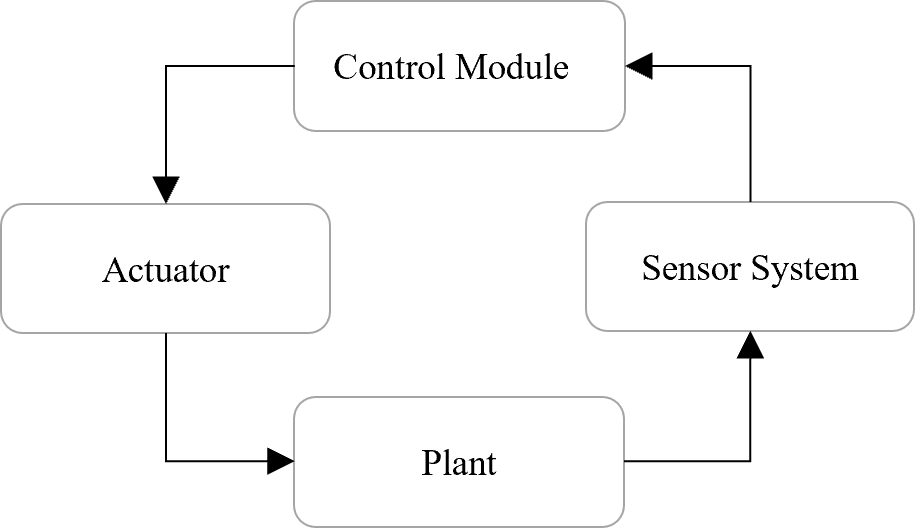}\label{fig:AEB_abstract_loop}}
  \hfill
  \subfloat[][Middle]
  {\includegraphics[width=0.33\textwidth]{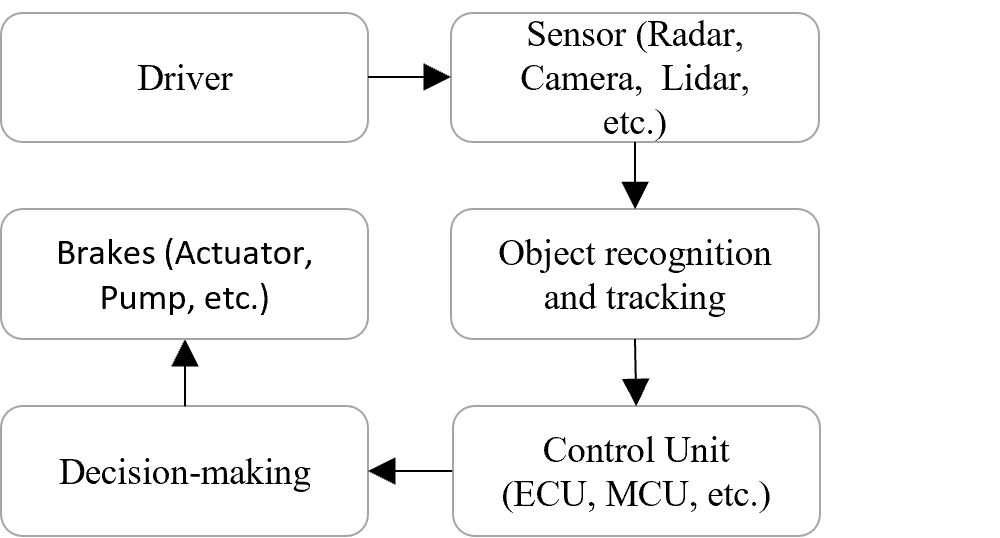}\label{fig:AEB_middle_loop}}
  \hfill
  \subfloat[][High]
  {\includegraphics[width=0.33\textwidth]{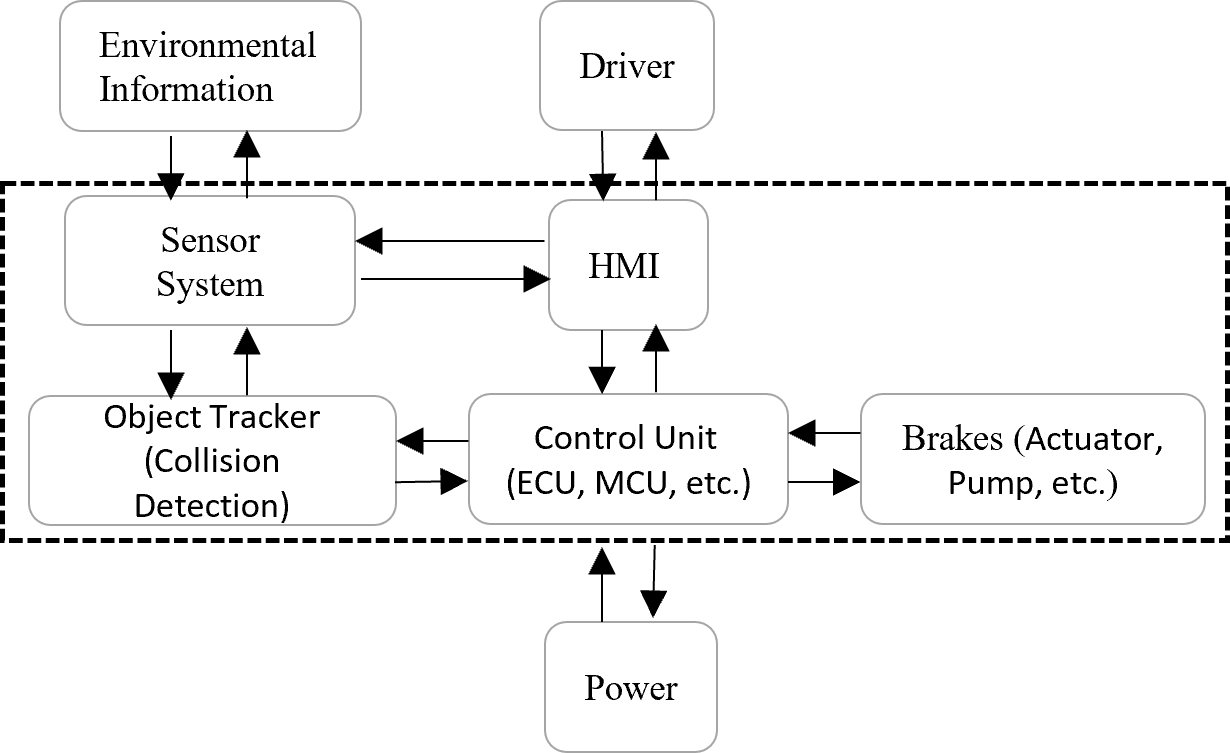}\label{fig:AEB_loop}}
  \hfill
  \\
  \subfloat[][Low]
  {\includegraphics[width=0.33\textwidth]{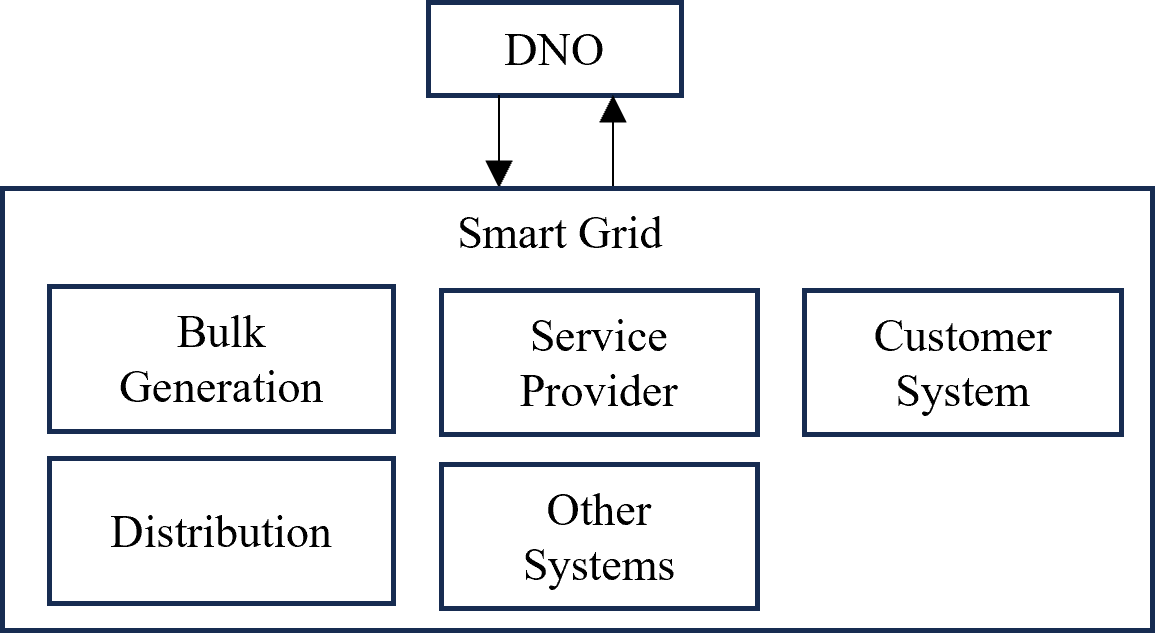}\label{fig:abstract_loop}}
  \hfill
  \subfloat[][Middle]
  {\includegraphics[width=0.33\textwidth]{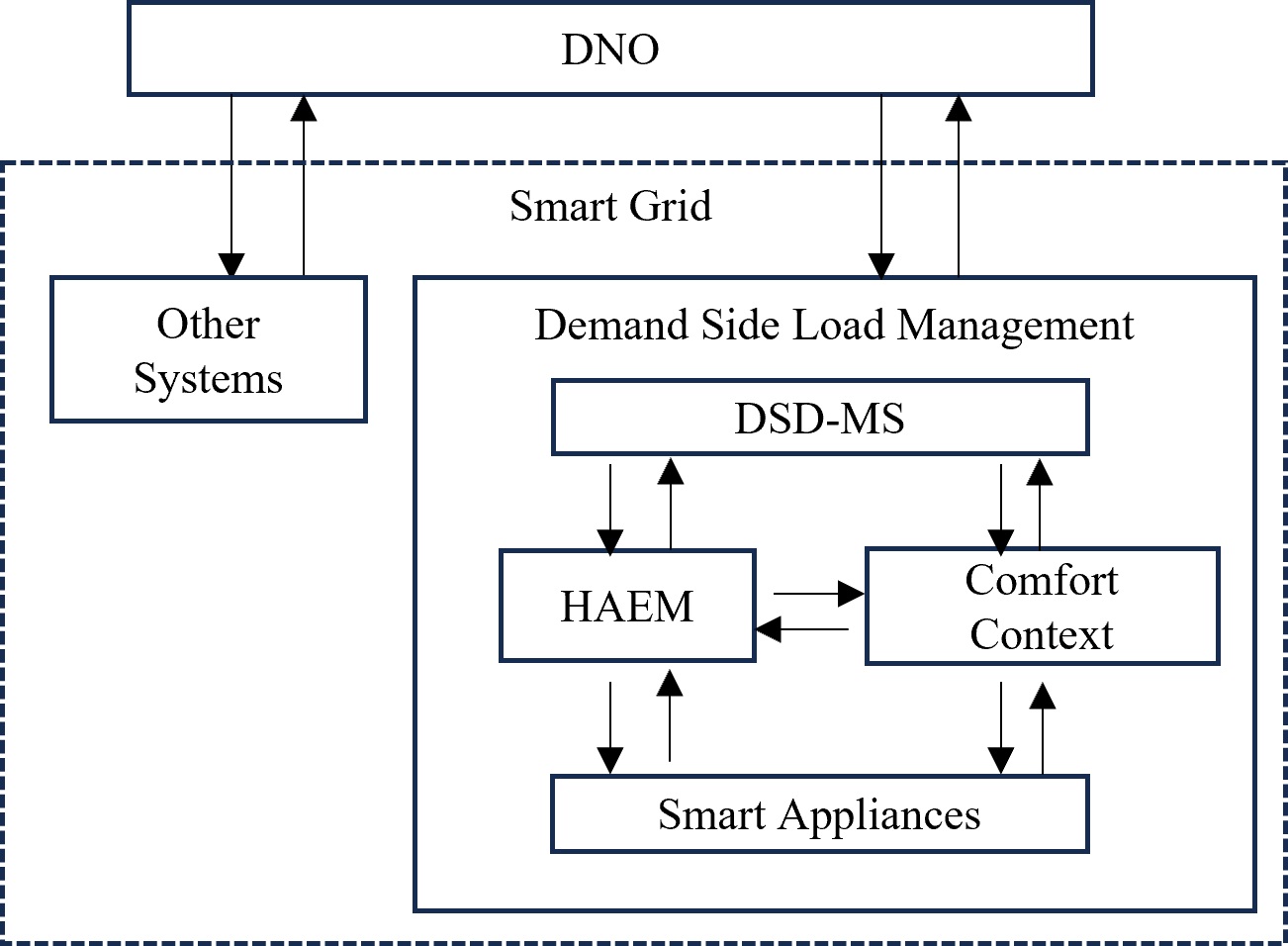}\label{fig:middle_loop}}
  \hfill
  \subfloat[][High]
  {\includegraphics[width=0.33\textwidth]{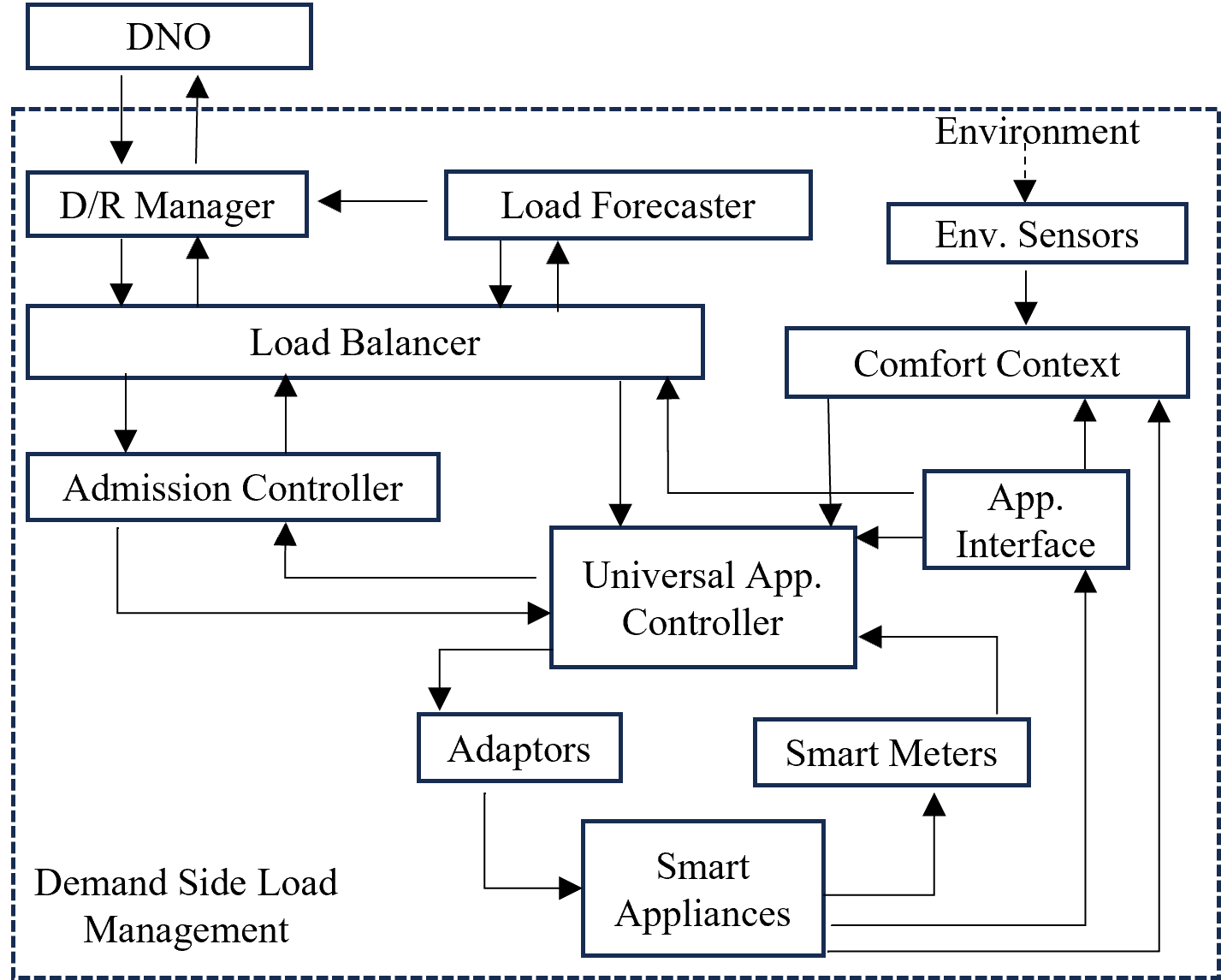}\label{fig:DSM_loop}}
  \caption{Control loop structures of three complexity levels for the two baselines, AEB (first row) and DSM (second row) systems.}
  \label{fig:baseline_loop_com}
\end{figure}

Fig.~\ref{fig:AEB_abstract_loop}, \ref{fig:AEB_middle_loop} and \ref{fig:AEB_loop} illustrates the three levels of complexity for the control loop structure of AEB systems, while Fig. \ref{fig:abstract_loop}, \ref{fig:middle_loop} and \ref{fig:DSM_loop} shows the three complexity variations for the control loop structure of DSM systems. The control loop structure for AEB systems was derived\footnote{ Following the methodology outlined in \cite{10.1115/1.4051940}, prompts were submitted to ChatGPT requesting generation of AEB system control loop diagrams at low, medium, and high levels of complexity.} from \cite{10.1115/1.4051940}. For DSM systems, the control loop structure diagrams of varying complexity were directly provided in \cite{karatzassystem}.

Each black arrow depicted in Fig.~\ref{fig:baseline_loop_com} denotes an individual control action, for which we ask ChatGPT to generate UCAs. That is, ChatGPT will be prompted to identify UCAs for each control action outlined in all 6 control loop structures in Fig.~\ref{fig:baseline_loop_com}. Each request made on a unique control action will be treated as an independent \textit{trial}, while each trial will generate a set of UCAs. Not all identified UCAs are correct\footnote{In this paper, we define correctness as an accurate and useful response that provides precise, credible details and reasoning to support any conclusions drawn. It contains factually correct information that directly answers the question or describes the topic at hand, without unrelated, false or illogical content.}. Thus, for each trial, we may record the following two metrics:
\begin{itemize}
    \item the \textit{quantity} of correct UCAs, which represents the \textit{comprehensiveness} of using ChatGPT in STPA; and
    \item the \textit{proportion} of correct UCAs out of the total UCAs identified, that represents the \textit{pertinence}.
\end{itemize}  
%The two selected metrics represents the \textit{comprehensiveness} and \textit{pertinence} of using ChatGPT in STPA, because: 
Becasue, a higher number of correct UCAs enables us to identify more causes of potential risks and hazards, which is instrumental in formulating more comprehensive safety constraints and requirements; Secondly, given ChatGPT's inherent propensity to generate noisy/irrelevant answers, it is important to check the proportion of correct UCAs in all UCAs generated.

% \yiqi{independent expert teams}
\paragraph{Evaluation Design}
\label{RQ2_eva_methods}
Since we will collect 3 groups (representing 3 control loop structures with different levels of complexity) of statistics on the two metrics---number and proportion of correct UCAs for each control action in Fig.~\ref{fig:baseline_loop_com}, we plan to conduct the Tukey-Kramer test, which is an established statistical technique for elucidating specific distinctions between groups. The test provides a range of values for each pairwise groups. If the difference between a pair of group means falls outside this range, then the difference is considered statistically significant \cite{richter2012using,abdi2010tukey}. Thus, we use the Tukey-Kramer test for testing the null hypothesis: \textit{The comprehensiveness and pertinence of UCA results do not significantly differ across groups when different complexity levels of control loop structures are modelled}.

% we plan to conduct the Analysis of Variance (ANOVA), which is an established statistical technique for analysing the differences in means between two or more groups. It operates by comparing the variance (a measure of variability) between the groups to the variance within each group. ANOVA allows to assess whether sample means from multiple groups have statistically meaningful distinctions \cite{st1989analysis, kim2017understanding}. 

\subsection{RQ3 Experiment and Evaluation Design}

\paragraph{Experiment Design}
%Additionally, the structure of the input prompts will be examined through establishing general guidelines for input formulation.
We aim to study the impact of using prompt guidelines in RQ3. Prompt guidelines is the skill of creating precise questions or cues that steer ChatGPT toward generating the answers. It serves as a critical link between what the user is looking for and what the model understands. Good prompts can substantially boost the quality and pertinence of ChatGPT's replies, whereas inadequate ones may yield incorrect or irrelevant outputs \cite{Ekin2023}. Normally, the standard prompt guide is as follows:
\begin{equation}
\label{eq_generic_prompt}
    Roles + Tasks + Steps + Limitation\; description + Output\; request
\end{equation}
For example, when ``\textit{Could you help me output a result using the STPA safety analysis method on an AEB system?}'' is an intended question for ChatGPT, the structured question using a prompt guide becomes: ``\textit{You are a safety expert, could you help me output a result using the STPA safety analysis method on an AEB system? Please provide more specific answers based on your previous responses, and avoid generating redundant answers.}'' 

As illustrated in the example, the provided template represents a general approach for posing questions to ChatGPT \cite{giray2023prompt}. However, its application in the domain of safety analysis, particularly concerning the characteristic of STPA, necessitates a specific focus. Thus, we design the the following STPA-specific prompt, which encapsulates the fundamental elements of UCAs (cf. expression \eqref{eq_uca_parts}):
\begin{equation}
\label{eq_stpa_prompt}
\begin{split}
    &Using\; STPA\; analysis,\; from <Source> send/recieve\; <Control Action> to <Controlled\; \\ &Component> in\; the <System>, list\; potential\; unsafe\; control\; actions, which\; includes\; all\; \\ &<Type>, <Context> and <Link\; to\; Hazards>
\end{split}
\end{equation}
For example, ``\textit{Using STPA analysis, from Didtribution Network Operator (DNO) send the request to the Smart Grid, list the potential unsafe control actions, which includes all type (provided, not provided, too early or too late, stopped too soon or applied too long), context and link to the Hazards.}'' is a question we built to query ChatGPT using the STPA-specific prompt guide.

In our later experiments for RQ3, the input prompts will be constructed in alignment with the aforementioned prompt guidelines. Each unique control action enumerated in Baseline \Rmnum{2} will be queried thrice---without the prompt guideline, with domain-agnostic prompt \eqref{eq_generic_prompt}, and with STPA-specific prompt \eqref{eq_stpa_prompt}. 
Similarly to RQ2, we opted for the third collaboration schemes and the third step of STPA, which involves identifying UCAs for each control action.

\paragraph{Evaluation Design}
% \xingyu{Since our statistical data is in the form of paired blah blah...}
% The independent samples t-test \xingyu{No? we are using paired sample t-test, aka *dependent* sampled t-test} is an inferential statistical analysis utilised to assess whether there exists a statistically significant difference between the mean values of two independent groups. It operates by comparing the actual difference between group means against the variation that would be expected due to random chance alone.  The t-test is an essential hypothesis testing tool for determining whether an independent variable has an effect on a continuous dependent variable between two experimental groups or populations \cite{Kalpić2011}. 

% We employ the Paired-Samples T Test for the statistical analysis of our data, which evaluates the mean difference between two measurements taken from the same individual, object, or related entity. These measurements, termed as ``paired'', are obtained under two distinct conditions. The objective of this test is to ascertain whether the mean difference between these paired observations deviates significantly from zero \cite{Kalpić2011,Kentspss}. That said, the Paired-Samples T-Test is well-suited for our purpose in analysing the paired statistical data on UCAs.

We collect a total of 38 control actions across the three control loop structures of Baseline \Rmnum{2} in Fig.~\ref{fig:baseline_loop_com}.
Thus, it generates 114 sets of UCAs in total after applying 3 different types of prompts for each control action. Similar to the evaluation design for RQ2, the current evaluation involves analysing the same control action across three conditions, thus the Tukey-Kramer test remains an appropriate statistical method for testing our null hypothesis: \textit{The comprehensiveness and pertinence of UCA results do not significantly differ across groups when various types of prompts are used}.
Again, we re-evaluate the two metrics in RQ2 to exam the comprehensiveness and pertinence, i.e., (i) the \textit{quantity} of correct UCAs and (ii) the \textit{proportion} of correct UCAs out of the total UCAs identified.

%Within the three defined complexity levels ins Baseline 2, each control action was queried. The UCAs corresponding to each control action were requested twice - first without utilising the prompt guide, second employing the systematic prompt guide. The quantity of correct UCAs was recorded for each control action, along with the proportion of correct UCAs out of the total UCAs generated. \xingyu{rephrase the move this part to the experiment design}

\section{Results}
\subsection{Answer to RQ1}

\begin{figure}[ht]
  \centering
   \subfloat[][Venn diagram of UCAs for AEB system]
  {\includegraphics[width=0.8\textwidth]{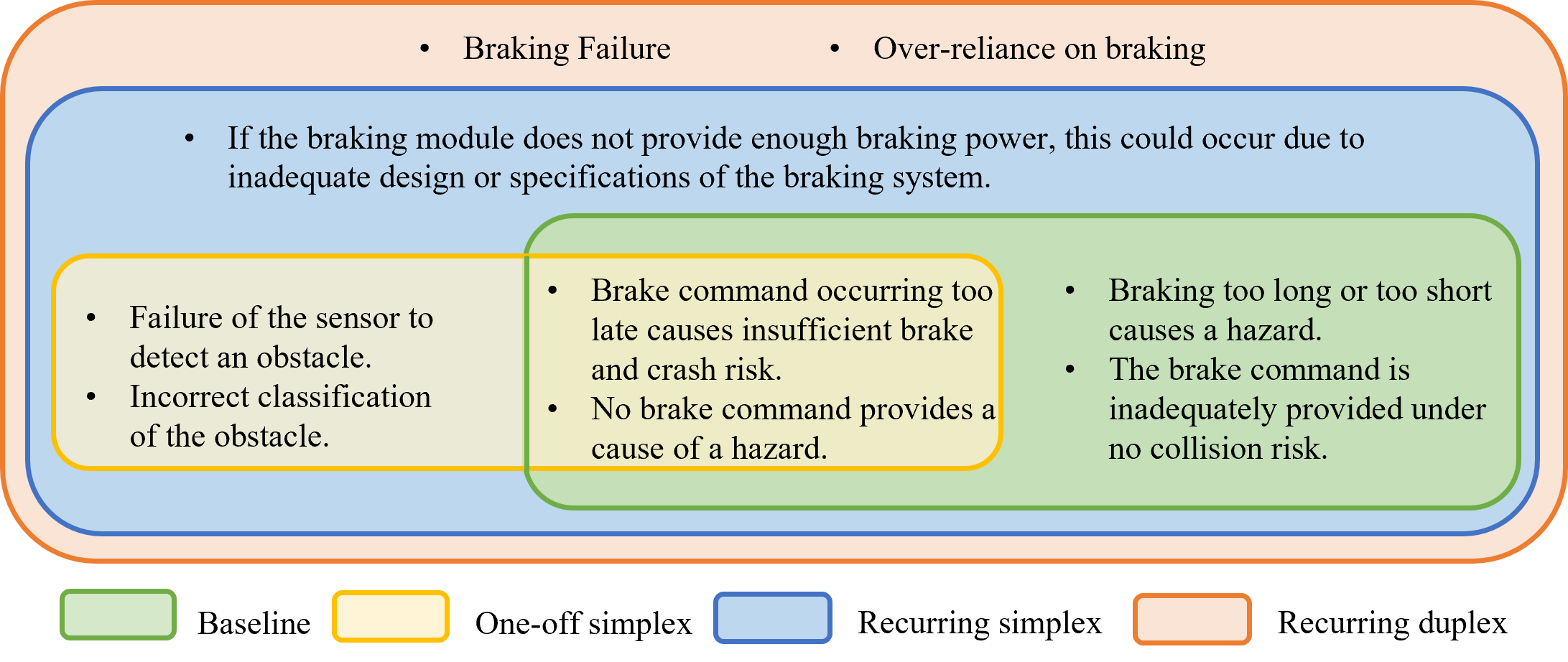}\label{fig8}}
  \hfill
  \subfloat[][Venn diagram of UCAs for DSM system]
  {\includegraphics[width=0.8\textwidth]{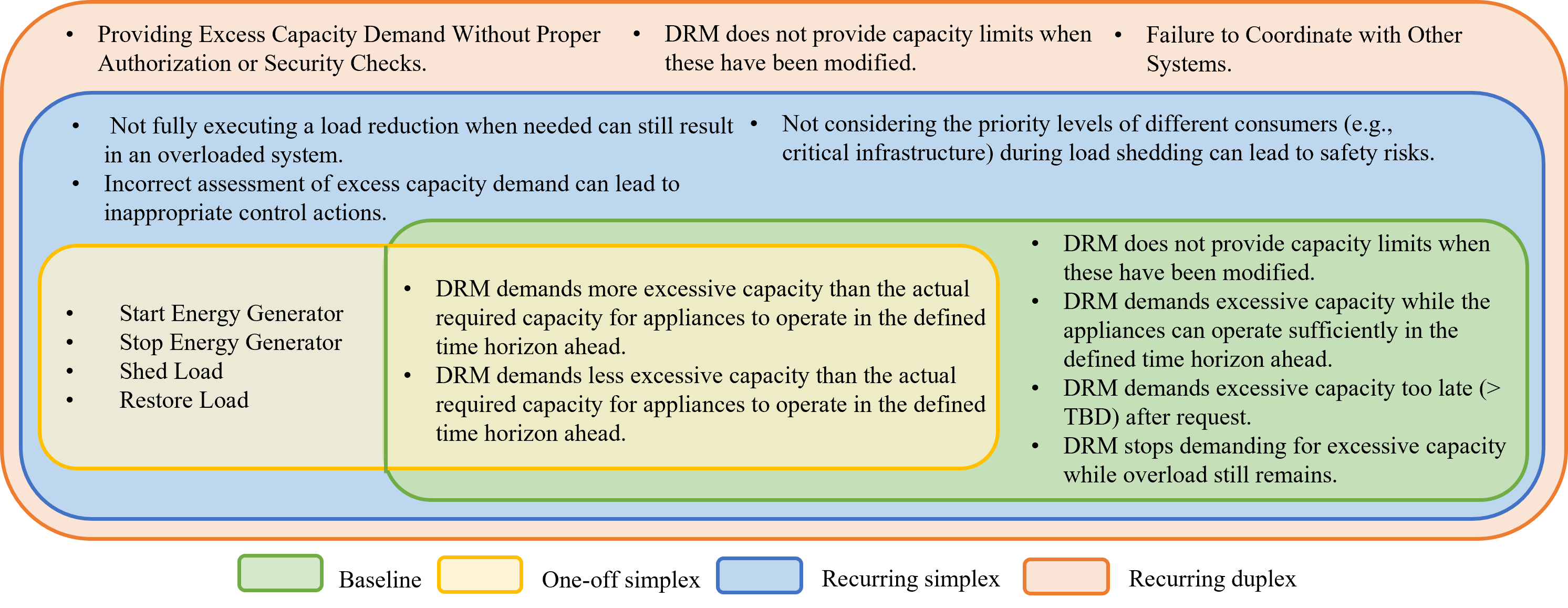}\label{rq1_dsm}}
  % \hfill
  \label{fig:baseline_loop}
  \caption{(a): The Venn diagram of the sets of UCAs for the AEB system. (b): The Venn diagram of the sets of UCAs for the DSM system. The different colour represents the baseline (green), \textit{one-off simplex collaboration case} (yellow), \textit{recurring simplex collaboration case} (blue) and \textit{recurring duplex collaboration case} (orange) respectively. \textbf{Note:} Owing to constraints pertaining to image dimensions, the UCAs has been selectively truncated to display only key elements. For the complete UCAs, cf. Appendix~\ref{app}
  % \yiqi{Legend scale}\xingyu{revise the figure according to the text, e.g. collaboration instead of interaction}
  }
\end{figure}

% This process of bias correction enhanced the accuracy and usefulness of the ChatGPT-generated answers. 
%To more effectively address RQ1, we conducted a comparative analysis of the UCAs identified across three interaction ways and baselines.\xingyu{???}
The comparative results to RQ1 can be found in Fig. \ref{fig8} and Fig. \ref{rq1_dsm}, and Table \ref{tab7}. The Venn diagrams in both Fig. \ref{fig8} and Fig. \ref{rq1_dsm} indicates: (1) although one-off simplex collaboration identifies new UCAs, it fails to identify all UCAs in the human baselines; (2) both recurring simplex and recurring duplex collaborations outperforms baselines by identifying a superset of UCAs, with recurring duplex collaboration being the best.

Based on the comparison attributes presented in Table \ref{tab3}, we have summarised the comparative study results of the three collaboration schemes in Table \ref{tab7}. 
% Recurring duplex collaboration stands out in terms of analysis results, while it necessitates/requires more domain knowledge from the safety experts, thus being complex/difficult and time-consuming\footnote{We have estimated the time cost of running each experiment in our studies.} than the other two collaboration schemes.\yiqi{Add some explanation for Table 4...}
The recurring duplex collaboration notably excels in the ``Analysis results'' aspect, achieving comprehensive identification of both hazards and causes. This contrasts with the partial identification achieved by both one-off simplex and recurring simplex schemes. However, this enhanced performance in the recurring duplex scheme comes with increased demands; it requires to handle more skills from safety experts, as indicated in the ``Analysis Process'' aspect. Furthermore, the recurring duplex collaboration is marked by a high complexity and difficulty level, which surpasses the low complexity associated with the other two schemes. In terms of time cost, while one-off simplex and recurring simplex collaborations are characterised by a low time investment, the recurring duplex scheme is, as expected, more time-consuming\footnote{Time cost is estimated from the time invested by the expert team in each interaction with the ChatGPT, in conjunction with the total number of interactions. A `One-off simplex' interaction typically encompasses 1 interaction, whereas a `Recurring simplex' interaction generally involves 5-6 interactions. Although the `Recurring duplex' interaction also entails 5-6 interaction, each interaction is further supplemented by a comprehensive analysis and organisation of individual steps by the expert team, culminating in the finalisation of results.}.

\begin{table}[h]
\caption{Comparative study results based on attributes listed in Table \ref{tab3}.}
\label{tab7}
\centering
\scalebox{0.9}{
\begin{tabular}{lllll}
\toprule
Aspects                     & Attributes          & One-off simplex       &Recurring simplex & Recurring duplex  \\ 
\midrule
Analysis results  & Identify hazards               &  Partially & Partially&Comprehensively  \\ 
Analysis results  & Identify causes               &  Partially & Partially & Comprehensively          \\ 
Analysis process  & Skills required                &  No&Few &Many          \\
Analysis process  & Time cost                 &  Low & Middle &High             \\ 
Analysis process  & Complexity/difficulty          & Low &Low &High\\
\bottomrule
\end{tabular}}
\end{table}

We note, the questions inputted to ChatGPT, along with their corresponding answers, are presented as \textit{\textbf{Q\&A}} blocks in the Appendix \ref{app}\footnote{For illustration, we only provide a partial list of Q\&A blocks while the complete results are available at our project website \url{https://github.com/YiQi0318/ChatGPT-STPA}\label{footnote1}}. A series of preliminary findings, including accuracy, robustness and reproducibility, identified and consolidated during the case studies are also noted as ``\textbf{Remarks}'' in the Appendix \ref{app}.

\subsection{Answer to RQ2}
\begin{figure}[htbp]
  \centering
  \subfloat[][Number of correct UCAs across 3 groups of samples]
  {\includegraphics[width=0.49\textwidth]{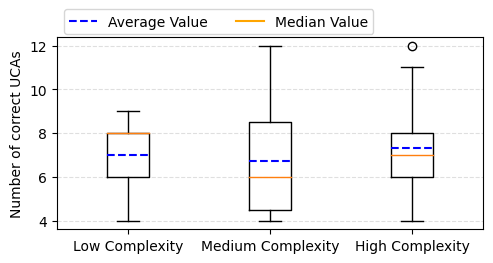}\label{fig:rq2_r22}}
  \hfill
  \subfloat[][Proportion of correct UCAs across 3 groups of samples]
  {\includegraphics[width=0.49\textwidth]{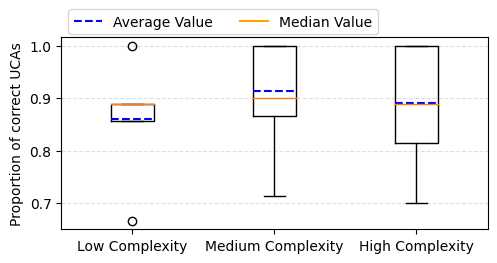}\label{fig:rq2_r2}
    }

  \caption{Box and whisker plots of samples for RQ2 
  % \xingyu{revise the figure according to the text..}
  }
  \label{fig:rq2_results}
\end{figure}

As per our experiment design of the RQ2, we took measurements of the following two metrics, the \textit{number} and \textit{proportion} of correct UCAs produced by ChatGPT for each control action. 
We conducted independent trials for all control actions from the two baseline systems and grouped them by the three levels of control loop complexity. The sample sizes for each group are: 5 for low complexity, 19 for medium complexity, and 30 for high complexity. While Fig.s \ref{fig:rq2_r22} and \ref{fig:rq2_r2} show the box and whisker plots of the two samples collected for the two metrics, we perform statistical analysis for the null hypothesis aforementioned.

% \xingyu{let us clarify ANOVA vs Tukey-Kramer test}

% \noindent
% \begin{minipage}{0.49\textwidth}
% \centering
% \begin{table}[H]
% \caption{Tukey-Kramer test results based on the number of correct UCAs}
% \label{tab_rq2_qu}
% \scalebox{0.7}{
% \begin{tabular}{lllllll}
% \toprule
% group1& group2& meandiff& p-adj &  lower&  upper&  reject \\
% \midrule
% Low   & Middle&   -0.2632& 0.9654& -2.7745& 2.2482&  False \\
% Low   & High  &   0.2 &0.9782 &-2.2135& 2.6135&  False \\
% Middle& High  &  0.4632& 0.7271& -1.0018& 1.9281&  False \\
% \bottomrule
% \end{tabular}
% }
% \end{table}
% \end{minipage}
% \hfill
% \begin{minipage}{0.49\textwidth}
% \centering
% \begin{table}[H]
% \caption{Tukey-Kramer test results based on the proportion of correct UCAs}
% \label{tab_rq2_prom}
% \scalebox{0.7}{
% \begin{tabular}{lllllll}
% \toprule
% group1& group2& meandiff& p-adj &  lower&  upper&  reject \\
% \midrule
% Low   & Middle&   0.0529& 0.5039& -0.0607& 0.1666&  False \\
% Low   & High  &   0.0136& 0.9518& -0.0957& 0.1228&  False \\
% Middle& High  &  -0.0394& 0.3318& -0.1057& 0.0269&  False \\
% \bottomrule
% \end{tabular}
% }
% \end{table}
% \end{minipage}
% \vspace{1em}

\begin{table}[h]
\centering
\begin{tabular}{llllllll}
\toprule
Metric&Group1& Group2& Meandiff& P-adj &  Lower&  Upper&  Reject \\
\midrule
\multirow{3}{*}{Number}&Low   & Middle&   -0.2632& 0.9668& -2.8306& 2.3042&  False \\
&Low   & High  &   0.3333 &0.9431 &-2.1341& 2.8007&  False \\
&Middle& High  &  0.5965& 0.6043& -0.9012& 2.0941&  False \\
\hline
\multirow{3}{*}{Proportion}&Low   & Middle&   0.0529& 0.5303& -0.0653& 0.1711&  False \\
&Low   & High  &   0.0294& 0.8070& -0.0842& 0.1431&  False \\
&Middle& High  &  -0.0235& 0.6910& -0.0925& 0.0455&  False \\
\bottomrule
\end{tabular}
\caption{RQ2: Tukey-Kramer test results based on the number of correct UCAs and the proportion of correct UCAs, when the system under analysis is modelled by control loop of different complexity}
\label{tab_rq2}
\end{table}

Table \ref{tab_rq2} depicts the Tukey-Kramer test results\footnote{Code and raw data are available at our project website \url{https://github.com/YiQi0318/ChatGPT-STPA}}. 
% The ``Meandiff'' columns represent the difference in means between the two groups being compared. A positive value indicates that the first group has a higher mean than the second group, while a negative value indicates the opposite. 
The ``Meandiff'' column represents the difference in means between the two groups being compared. A positive value indicates that the mean of the second group is higher than that of the first group, while a negative value indicates that the mean of the second group is lower.
The $p$ value is a measure used to determine statistical significance in hypothesis testing. 
The ``adjusted'' p-value (denoted as ``P-adj'') is a corrected p-value that accounts for multiple comparisons. We may observe that all group-wise comparisons in the tables have a P-adj greater than the common significance threshold $\alpha=0.05$, 
% indicating the acceptance of the null hypothesis,
fail to reject the null hypothesis as shown in the last columns of the table.
``Lower'' and ``Upper'' means the lower bound and upper bound of the confidence interval. The calculation of these bounds is based on the standard error of the estimate and a critical value from a relevant statistical distribution. Specifically, it is obtained by adding or subtracting the margin of error from the ``Meandiff''. These bounds provide a range within which we can expect the true mean difference to lie with the specified level of confidence. A narrower interval indicates a more precise estimate, while a wider interval suggests greater uncertainty.

\subsection{Answer to RQ3}

\begin{figure}[htbp]
  \centering

  \subfloat[][Number of correct UCAs across 3 groups of samples]
  {\includegraphics[width=0.49\textwidth]{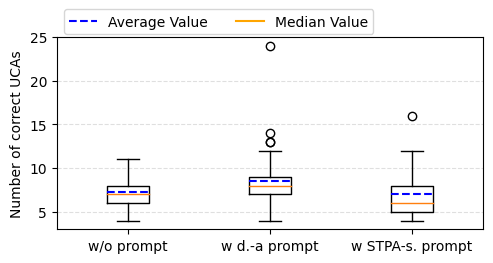}\label{fig:rq3_r2}}
  \hfill
    \subfloat[][Proportion of correct UCAs across 3 groups of samples]
  {\includegraphics[width=0.49\textwidth]{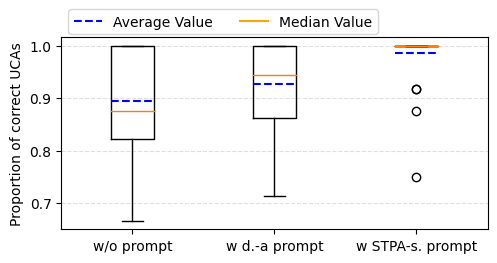}\label{fig:rq3_r22}}
  \caption{Box and whisker plots of samples for RQ3 (NB: ``d.-a.'' refers to ``domain-agnostic'', ``STPA-s.'' refers to ``STPA-specific''). 
  % \xingyu{Please update the fig using d.-a. prompt etc, and omit the ``input" maybe; consider add in a new line showing the mean for each group}
  % \xingyu{revise like the fig above. change the order, rename the caption}
  }
  \label{fig:rq3_results}
\end{figure}

We identified a total of 38 control actions within Baseline \Rmnum{2} across its three control loop structures. For each control action, we queried ChatGPT thrice, as per the experiment design. The box plots in Fig.~\ref{fig:rq3_results} visualise the samples, based on which we employed the Tukey-Kramer test to assess the statistical outcomes. Again, our null hypothesis is: There is no significant difference between the groups. Detailed results are presented in Table \ref{tab_rq3}.

% \begin{table}[h]
% \centering
% \caption{Paired t-test results based on the number of correct UCAs and the proportion of correct UCAs}
% \label{tab_rq3}
% \scalebox{0.9}{
% \begin{tabular}{llllllll}
% \toprule
% Metric &Group1   & Group2   &   Meandiff &     P-adj &        Lower &     Upper &   Reject \\
% \midrule
% Number & w prompt  & w/o prompt  &  1.26316 & 0.0514498 & 0.185377 & 2.34094 &        False 
% \\
% Proportion & w prompt  & w/o prompt  &  0.0325101 & 0.0960405 & -0.00541903 & 0.0704392 &        False \\
% \bottomrule
% \end{tabular}
% }
% \end{table}
\begin{table}[h]
\centering
\scalebox{0.95}{
\begin{tabular}{llllllll}
\toprule
Metric&Group1& Group2& Meandiff& P-adj &  Lower&  Upper&  Reject \\
\midrule
\multirow{3}{*}{Number}&w/o prompt   & w d.-a. prompt&   1.2632& 0.1057& -0.2021&  2.7284&  False \\
&w/o prompt   & w STPA-s. prompt  &   -0.2368&  0.922& -1.7021&  1.2284&  False \\
&w d.-a. prompt& w STPA-s. prompt  &  -1.5000& 0.0435& -2.9653& -0.0347&   \textbf{True} \\
\hline
\multirow{3}{*}{Proportion}&w/o prompt   & w d.-a. prompt&   0.0325& 0.1662& -0.0098& 0.0748&  False \\
&w/o prompt   & w STPA-s. prompt  & 0.0922&    0.0000 &  0.0499& 0.1346&   \textbf{True} \\
&w d.-a. prompt& w STPA-s. prompt  &  0.0597& 0.0031&  0.0174&  0.102&   \textbf{True} \\
\bottomrule
\end{tabular}}
\caption{RQ3: Tukey-Kramer test results based on the number of correct UCAs and the proportion of correct UCAs, when using different types of prompts (NB: ``d.-a.'' refers to ``domain-agnostic'', ``STPA-s.'' refers to ``STPA-specific'').}
\label{tab_rq3}
\end{table}

Regarding the comprehensiveness of correct UCAs identified, we only observe that using domain-agnostic prompts can significantly identify more UCAs than using STPA-specific prompts. This finding is counter-intuitive, as one might expect domain-specific prompts to be more effective. We hypothesise that the constraints of STPA-specific prompts lead ChatGPT to generate more ``conservative'' responses. This suggests a potential limitation in the scope of answers generated by more narrowly focused prompts. 
% \begin{remark}
% \label{remark_rq3_1}
% \end{remark}

On the other hand, using STPA-specific prompts can significantly enhance the pertinence of ChatGPT's answers compared to using either domain-agnostic prompts or no prompts at all. This supports our earlier hypothesis that ChatGPT, when guided by more specific prompts, tends to be more conservative, identifying fewer UCAs in an effort to ensure the correctness of those identified.

\subsection{Threats to Validity}
%In this subsection, ``validity'' denotes the precision of our findings. Our examination of the particular comparative tasks' validity is grounded in a case study framework, encompassing three dimensions: construct validity, internal validity, and external validity \cite{runeson2009guidelines}.\xingyu{Not sure we need this paragraph at all...}

\paragraph{Construct Validity}
``Construct validity'' pertains to the degree to which the investigative goals and outcomes accurately measure their targeted constructs. It assesses whether a test or measurement tool truly measures what it claims to measure.
In our study, we collect data regarding the correct UCAs to do both qualitative and quantitative analyses. But the determination of ``correctness" of a UCA is prone to bias, as it depends on the specific background and expertise of the human experts involved. To mitigate, we engaged two independent teams of experts to validate the case study results. We also make all intermediate results publicly accessible on our project website for feedbacks to further mitigate this threat.
In RQ3, we aim to examine the effect of using prompt guidelines, for this purpose, we employed both a domain-agnostic guide and a STPA-specific prompt guide. However, only one STPA-specific prompt guide was used, which poses a potential threat. We recognise that prompt engineering is a rapidly evolving field, potentially leading to new methods for crafting more effective domain-specific prompts. In future, we plan to diversify the STPA-specific prompt guidelines in our study to mitigate this threat further.

%Selecting an overly simplistic system might mask disparities in the outcomes derived from different analytical approaches. For example, since the sample size within each individual baseline case was insufficient, we made the methodological decision to combine two baselines case together. However, even after aggregation, the total sample size remained below common minimum recommendations for robust statistical analysis. Small sample sizes are acknowledged to increase the potential for outliers and negatively influence overall results. With fewer data points, a few anomalous values can skew measures of central tendency and distribution shape. Consequently, outliers may erroneously appear more extreme and impactful than they would be in larger samples. The small sample size also reduces statistical power to detect true effects. The presence of potential outliers and reduced reliability must be interpreted conservatively in light of the constraints around sample size. Any conclusions or inferences drawn from analyses should factor in these inherent limitations and the potential for inflated error effects. 

\paragraph{Internal Validity}

Threats may correspond to bias in establishing cause-effect relationships in our case study. In RQ2 and RQ3, we conducted well established statistical testing techniques to draw the causal-effect conclusions. Confounding variables, i.e., factors other than the control loop structure complexity and prompt guideline (e.g., tone and politeness) may have influenced outcomes of correct UCAs. To mitigate such threat, we plan to do more Randomised Controlled Trials in the future by carefully considering and addressing confounding variables in the design and analysis. Moreover, when making claims on the effectiveness of using ChatGPT in STPA, our focus has been on experimenting with the step of finding UCAs in STPA. Although it is arguably the key step of STPA, this threat can be mitigated by completing the experiments on other STPA steps in future.

%``Internal validity'' is predominantly influenced by the meticulousness of the study's methodologies and their execution \cite{Bhandari_2022b}. Potential pitfalls encompass the analysts' subjective influence on the case studies. Particularly potential discrepancies might arise if the expert teams lacks comprehensive familiarity with the systems under scrutiny. In the paper, the members of expert teams collaboratively reviewed and interpreted the results obtained from the STPA methodology. As with any analysis conducted by humans, inherent biases may shape the findings in subtle ways. The backgrounds, prior experience, and perspectives of the experts conducting the analysis can unconsciously influence what results are focused on and how they are synthesised. 

\paragraph{External Validity} 
Factors limiting generalisability threaten external validity. In this regard, we identify three potential threats: (1) the relatively small sample size in RQ2 and RQ3; (2) only two baseline systems (published in peer-reviewed works) were selected across the 3 RQs; (3) only one ChatGPT version was studied. To mitigate them, more sample needs to be collect with diversified data representing more systems under study and LLMs. In this preliminary study, we have explicitly discussed the limitations related to our sample’s representatives.

%``External validity'' refers to the universality of research results \cite{Bhandari_2022b}. Intuitively, one may hypothesise that employing systematic prompt guidelines would improve results. However, the experimental findings indicated no statistically significant differences in outputs when utilising the prompt framework versus standard prompts. While the prompt guidelines did yield more responses overall, they did not increase the proportion of accurate options generated. This raises questions regarding the necessity of domain-specific customisation for prompt guidelines to be effective. It's imperative to acknowledge that the foundational models of LLMs in prevailing applications are perpetually evolving. For instance, within this paper, ChatGPT transitioned from GPT3.5 to GPT4. Such advancements pose significant challenges to the generalisability of research findings and notably impact the external validity.

\section{Discussion and Conclusion}

\subsection{Expanding the Research Horizon}
We pose the following questions to initiate the discussion of future research in this direction.

\paragraph{What is the top concern of using LLMs in safety analysis, and how can it be addressed?} LLMs require a large amount of high-quality data for effective training, however, safety-critical industries may not have enough data available for this purpose, which may lead to LLMs not being able to output specific details of the safety analysis domain but only general information. There is currently no widely accepted standard for evaluating the accuracy and robustness of LLMs' results. Additionally, the system may be vulnerable to unauthorised data disclosure and privacy violations, as well as backdoor attacks that bypass normal authentication \cite{khowaja2023chatgpt,huang2023survey}. Having said all that, we believe the top concern of using LLMs for safety analysis is the trustworthiness issues.
\textit{To make the safety analysis results trustworthy, a range of LLMs' properties need to be considered, such as accuracy, reliability, generalisation, robustness, interpretability, and fairness} \cite{openai2023gpt4,zhao2023survey}. With advancements in LLMs technology and increased research efforts in the field, issues surrounding the trustworthiness of LLMs may be resolved to a certain extent, such as the recent efforts \cite{wang2023robustness,perez2022red}.

\paragraph{How does the rapid and constant updating of LLMs impact safety analysis?}
The field of LLMs is currently undergoing rapid development, and the frequent updates and iterations of models like GPT-4 (published in March 2023) can contribute to lowering the blue bar in Fig.~\ref{fig4}, indicating an improvement in the generlisability of LLMs and a broader range of risks that can be addressed. Traditional safety assurance processes are static and carried out prior to the system's deployment \cite{calinescu_engineering_2018}, but the emergence of \textit{dynamic assurance} presents an ongoing metric for continuously assessing safety assurance to ensure ongoing safety \cite{9269875,calinescu_engineering_2018}.
\textit{The dynamic and constantly evolving nature of LLMs is well-suited for the dynamic assurance concept in safety analysis, while how to integrate them in a principled way require more investigation}.
%Furthermore, integrating \gls{LLMs} with current safety analysis process can be challenging due to differences in terminology and methodologies. This is why the development of the three use cases presented in this paper was intended to address this issue.

%LLMs also shares some common weaknesses, including potential reasoning errors, potential security vulnerabilities and the datasets used by LLMs cannot incorporate the most recent data.

\paragraph{What about other types of safety analyses?} In this paper, we have opted to utilise the STPA method because of its popularity \cite{harkleroad2013review,leveson2018stpa}. For future research, \textit{it is worthwhile to investigate other safety analysis methods, e.g. HAZOP, FTA and FMEA, as well, and determine which safety analysis method can be more effectively combined with LLMs}, and how to do so.

\paragraph{How to facilitate the practical and large-scale adoption of LLMs in safety analysis?} We believe the key step is standardisation, which is a pressing need in this emerging approach to safety analysis. All aspects of integrating LLMs into the safety analysis process (e.g., terminologies, actives, evaluations metrics, etc.) should be agreed upon, formalised, documented in safety standards, and subject to regulatory oversight.

\subsection{Conclusion}
Impressed by the recent successful stories of ChatGPT in many domains, we pose the question: ``if safety analysis can actually make use of LLMs?''. To answer, we conducted a case study of applying ChatGPT in the STPA for a AEB system and a DSM system, which is the first attempt of using LLMs in safety analysis, to the best of our knowledge.

Our findings centre around the effectiveness, efficiency, and relevance of STPA when using various collaboration approaches with ChatGPT. These claims are supported by both comparative studies and statistical analyses, highlighting the potential of LLMs in addressing the inherent challenges of high complexity and subjectivity in STPA. Our work contributes by offering preliminary yet valuable insights toward the development of a more intelligent safety analysis framework that leverages the capabilities of LLMs.

%identify the most effective way of incorporating ChatGPT into the STPA process, which may address the high complicity and subjectivity limitations of STPA. While we concluded remarks and research questions representing the opportunities and challenges, 

% \section*{Acknowledgments}
% This work is supported the U.K. EPSRC through End-to-End Conceptual Guarding of Neural Architectures [EP/T026995/1]. This project has received funding from the European Union’s Horizon 2020 research and innovation programme under grant agreement No 956123. YQ’s contribution to the work is supported through Chinese Scholarship Council. \xingyu{To thank the two safety expert teams; to add back later}

\section*{Data availability}
All data generated in this project is publicly available at \url{https://github.com/YiQi0318/ChatGPT-STPA}.

\section*{Code availability}
All code developed in this project is publicly available at \url{https://github.com/YiQi0318/ChatGPT-STPA}.

% \section{Author contributions}

% \noindent QY: Conceptualisation, data curation, experimentation, result evaluation and analysis, validation, writing;

% \noindent XZ: Conceptualisation, data curation, experimentation, result evaluation and analysis, project administration, supervision, validation, writing;

% \noindent SK: Result evaluation and analysis, project administration, validation, writing;

% \noindent XH: Conceptualisation, result evaluation and analysis, funding acquisition, project administration, supervision, validation, writing;

% \section*{Competing interests}
% The authors declare no competing interests.

\bibliographystyle{plain}
\bibliography{ref}

\clearpage
\appendix

\section{Q\&A Examples with ChatGPT and Remarks}
\label{app}

\subsection{One-off simplex collaboration}
The first case shown in Fig.~\ref{fig2}(a) involves the expert team providing ChatGPT with the keywords ``AEB System" and ``STPA Method", without sharing the specific details of how to perform \gls{STPA}. Then, STPA is conducted automatically by ChatGPT, according to the steps obtained directly from ChatGPT. This case is referred as ``One-off simplex collaboration'', in the sense that the only collaboration between human experts and ChatGPT is at the very beginning and the communication diction is single-directional from human to ChatGPT. Intuitively, this case represents the scenario in which the human experts fully trust ChatGPT and only need to specify their high-level inputs (i.e., conducting STPA on the AEB system or DSM system) to get the final analysis results from STPA.

To start, we may conduct an initial query to ChatGPT regarding the fundamental process of \gls{STPA} and how to apply it to the \gls{AEB} system.

\begin{tcolorbox}[enhanced,breakable,title = {AEB Example Q\&A 1.1}]
	\textbf{Q:} \textit{How to use STPA to analyse AEB systems?}
	\tcblower
	\textbf{A:} 
	(a) Define the system boundaries;
	(b) Identify the top-level hazards;
	(c) Develop control structure diagrams;
	(d) Identify potential unsafe control actions;
	(e) Analyze the causes of the unsafe control actions;
	(f) Develop recommendations to mitigate hazards.
\end{tcolorbox}

\begin{tcolorbox}[enhanced,breakable,title = {DSM Example Q\&A 1.1}]
	\textbf{Q:} \textit{How to use STPA to analyse Electricity Demand Side Management (DSM) systems?}
	\tcblower
	\textbf{A:} 
	(a) Step 1: System Description;
	(b) Step 2: Identify Unsafe Control Actions;
	(c) Step 3: Analyze Control Actions;
	(d) Step 4: Identify Contributing Factors;
	(e) Step 5: Generate Safety Requirements;
	(f) Step 6: Generate Recommendations;
        (g) Step 7: Validate and Iterate.
\end{tcolorbox}

The original \gls{STPA} process consists of five basic steps, cf. Section \ref{sec_stpa}. However, in \textit{AEB Example Q\&A 1.1} (cf. the Appendix), ChatGPT outputs eight steps. Despite the discrepancy, we can perform STPA by following the the six steps which arguably is a more refined version of the original five-step STPA. In the \textit{DSM Example Q\&A 1.1} from baseline \Rmnum{2}, we also observe that ChatGPT indicates STPA necessitates a full seven-step analysis.

\begin{remark}[Accuracy despite discrepancy]
\label{rm_diff}
Outputs from ChatGPT regarding the steps of conducting STPA may be different from those in published books and academic papers, but not necessarily incorrect.
\end{remark}

\begin{tcolorbox}[enhanced,breakable,title = {Example Q\&A 1.2A}]
	\textbf{Q:} \textit{Could you help me output a result using STPA safety analysis method on an AEB system?}
	\tcblower
	\textbf{A:} %\\
	(a) System Boundaries: Vehicle, AEB system, Environment.
	
	(b)Top-Level Hazards:
	1. Collisions with other vehicles
	2. Collisions with pedestrians
	3. Collisions with objects
	
	(c) Control Structure Diagram:
	1. Sensors (camera, radar, lidar)
	2. Software (object detection, decision-making, braking control)
	3. Braking system
	
	(d) Potential Unsafe Control Actions:
	1. Failure of the sensor to detect an obstacle
	2. Incorrect classification of the obstacle
	3. Delay in the decision-making process
	4. Failure of the braking system to activate
	
	(e) Causes of Unsafe Control Actions:
	1. Sensor malfunction due to hardware or software failure
	2. Insufficient training data for the object detection software 
	3. Environmental factors such as weather conditions or lighting that interfere with sensor accuracy
	4. Software bugs that cause delays or incorrect classification of obstacles
	5. Malfunction of the braking system due to hardware failure or lack of maintenance
	
	(f) Recommendations to Mitigate Hazards:
	1. Improve sensor redundancy and fault tolerance
	2. Increase training data for the object detection software to improve accuracy
	3. Develop algorithms to adjust for environmental factors that can impact sensor accuracy
	4. Implement thorough testing procedures to detect and address software bugs
	5. Implement a preventative maintenance program for the braking system to ensure proper function and reduce the risk of failure.
\end{tcolorbox}

\begin{tcolorbox}[enhanced,breakable,title = {Example Q\&A 1.2B}]
	\textbf{Q:} \textit{Could you help me output a result using STPA safety analysis method on a Electricity Demand Side Management (DSM) system?}
	\tcblower
	\textbf{A:} %\\
	Step 1: System Description
        Functional Model:
            System Components: DSM Controller, Energy Storage, Energy Generator, Consumers
            Functional Description: The DSM controller issues commands to manage energy distribution based on current demand and supply status.
        Control Structures:
            DSM Controller to Energy Storage: Issue charge or discharge commands
            DSM Controller to Energy Generator: Issue start/stop generation commands
            DSM Controller to Consumers: Issue load-shedding or load-restoration commands
        Control Actions:
            DSM Controller to Energy Storage: [Charge, Discharge]
            DSM Controller to Energy Generator: [Start, Stop]
            DSM Controller to Consumers: [Shed Load, Restore Load]
    Step 2: Identify Unsafe Control Actions
        List of Hazardous States:
            Overloading of the energy generator
            Energy storage failure due to overcharging or excessive discharging
            Consumer dissatisfaction due to unnecessary load shedding
        Map Control Actions to Hazards:
            Charge Energy Storage: Could lead to energy storage failure if overcharged
            Discharge Energy Storage: Could lead to energy storage failure if excessively discharged
            Start Energy Generator: Could lead to generator overloading
            Stop Energy Generator: Could lead to unmet consumer demand
            Shed Load: Could lead to consumer dissatisfaction
            Restore Load: Could lead to generator overloading if not properly managed
    Step 3: Analyse Control Actions
        Timing Conditions:
            Charge Energy Storage: Hazardous if initiated when storage is nearly full
            Discharge Energy Storage: Hazardous if initiated when storage is nearly empty
            Start Energy Generator: Hazardous if initiated when already at capacity
            Stop Energy Generator: Hazardous if initiated during peak demand
            Shed Load: Hazardous if initiated during non-critical periods
            Restore Load: Hazardous if initiated without adequate generation capacity
        Omissions and Commissions:
            Failure to Shed Load: Could lead to generator overloading
            Unnecessary Load Shedding: Could lead to consumer dissatisfaction
            Failure to Charge Storage: Could lead to unmet demand during peak hours
    Step 4: Identify Contributing Factors
        Human Errors: Misinterpretation of energy supply and demand data
        System Failures: Malfunctioning sensors or actuators
        External Factors: Cyber-attacks, extreme weather conditions
    Step 5: Generate Safety Requirements
        The system must prevent energy storage from overcharging or excessively discharging.
        The system must not overload the energy generator.
        The system must minimise consumer dissatisfaction due to load shedding.
    Step 6: Generate Recommendations
        Implement safety limits on the charge and discharge levels of energy storage.
        Implement real-time monitoring of the energy generator and automatically adjust operations to prevent overloading.
        Implement a consumer priority system for more intelligent load shedding.
    Step 7: Validate and Iterate
        Incorporate safety requirements into the system design and operation.
        Perform tests to validate that the safety requirements are met.
        Continuously monitor the system for emerging hazards and update the safety analysis as needed.

\end{tcolorbox}

In \textit{AEB Example Q\&A 1.2}, ``Failure of the sensor to detect an obstacle'' and ``Incorrect classification of the obstacle'' correspond to ``No brake command provides a cause of a hazard'' in the baseline, and ``Delay in the decision-making process'' may correspond to the ``Brake command occurring too late causes insufficient brake and crash risk''.

\begin{remark}[Unreliability]
\label{rm_unreli}
    The STPA result of \textit{One-off simplex collaboration case} is inconsistent with the baseline by human safety experts, as the UCAs discovered is only a subset of those in Table \ref{tab: accidents_aeb}. Specifically, ``The brake command is inadequately provided under no collision risk'' and ``Braking too long or too short causes a hazard'' are not identified in \textit{One-off simplex collaboration case}. 
    Therefore, it can be remarked that \textit{One-off simplex collaboration case} is unreliable in terms of missing important UCAs, compared to the baseline.
\end{remark}

\begin{remark} [Propagation and compounding of errors]
\label{rm_propa}
    From \textit{AEB Example Q\&A 1.1} to \textit{AEB Example Q\&A 1.2}, it suggests that, an error made in initial steps can cause greater deviation in the subsequent steps in the end-to-end process of STPA without any human interventions.
\end{remark}

By \textit{AEB Example Q\&A 1.1} and \textit{AEB Example Q\&A 1.2}, we have demonstrated the complete process using the one-off simplex collaboration. However, for evaluation purpose, we also want to get intermediate results from ChatGPT in this case for later comparative studies. Thus, we make an additional query to ChatGPT on the control loop structure in the following \textit{AEB Example Q\&A 1.3}.

\begin{tcolorbox}[enhanced,breakable,title = {Example Q\&A 1.3A}]
	\textbf{Q:} \textit{Could you help me draw the model control loop structure on the AEB systems?}
	\tcblower
	\textbf{A:} %\\
	(a) Sensor input: The system takes in data from various sensors such as radar, lidar, and cameras to detect potential obstacles or hazards on the road.
	(b) Object recognition and tracking: The system uses algorithms to analyze the sensor data and identify objects such as vehicles, pedestrians, and cyclists. It then tracks the movement of these objects to determine if they pose a risk of collision.
	(c) Decision-making: Based on the sensor data and object recognition/tracking, the system makes a decision on whether to apply the brakes or not. If the system determines that a collision is imminent, it will activate the braking system.
	(d) Actuation: Once the decision is made to apply the brakes, the system sends a signal to the braking system to engage the brakes and stop the vehicle.
	(e) Feedback: The system continuously monitors the situation and provides feedback to the driver, such as visual and audible warnings, to alert them to the potential danger.
\end{tcolorbox}

\begin{tcolorbox}[enhanced,breakable,title = {Example Q\&A 1.3B}]
	\textbf{Q:} \textit{Could you help me draw the model control loop structure on the Electricity DSM systems?}
	\tcblower
	\textbf{A:} %\\
	Elements:
            Controllers: DSM Controller, Consumer Home Automation Systems
            Controlled Processes: Energy Storage, Energy Generator, Consumer Appliances
            Actuators: Charging Unit, Generator Control System, Load Shedding Systems
            Sensors: Energy Level Monitor, Generator Status Monitor, Consumer Feedback Monitor
        Control Loops:
            DSM Controller to Energy Storage
            Controller: DSM Controller
            Controlled Process: Energy Storage
            Actuator: Charging Unit
            Sensor: Energy Level Monitor
            Control Actions: Charge, Discharge
            Feedback Loop: Energy level monitored by Energy Level Monitor sent back to DSM Controller
        DSM Controller to Energy Generator
            Controller: DSM Controller
            Controlled Process: Energy Generator
            Actuator: Generator Control System
            Sensor: Generator Status Monitor
            Control Actions: Start, Stop
            Feedback Loop: Generator status monitored by Generator Status Monitor sent back to DSM Controller
        DSM Controller to Consumers
            Controller: Consumer Home Automation Systems
            Controlled Process: Consumer Appliances (e.g., HVAC, lights)
            Actuator: Load Shedding Systems
            Sensor: Consumer Feedback Monitor
            Control Actions: Shed Load, Restore Load
            Feedback Loop: Consumer feedback monitored by Consumer Feedback Monitor sent back to Home Automation Systems, which may also communicate with DSM Controller
\end{tcolorbox}

Unfortunately, as a language model, ChatGPT is unable to create visual diagrams. Instead, it provides the expert team a text guidance for creating graphical outputs. In \textit{DSM Example Q\&A 1.2}, it's evident that when ChatGPT fails to fully comprehend the preceding question, the subsequent answers diverge significantly from those provided by expert knowledge (see Remark \ref{rm_propa}). The discrepancy in results (experts results as shown in Table \ref{tab: accidents}) is largely attributable to ChatGPT's inability to process images; it can only interpret keywords or textual descriptions of the system's structure (as shown in Remark \ref{rm_no_graph}).
Different users may interpret text-based guidance differently, resulting in varying forms of the control loop structure. For instance, it may be unclear how to properly draw feedback arrows.
Typically, traditional STPA focuses on identifying high-level consequences, such as traffic accidents or rear-end collisions, during its initial stage. When using ChatGPT for STPA analysis, the situation might differ. If you pose the same question to ChatGPT multiple times, be aware that the answers could vary.

\begin{remark}[Graphical outputs]
\label{rm_no_graph}
As expected, the current version of ChatGPT cannot generate graphic outputs, while the text-based description provides guidance on how to create the diagrams. 
That said, it is also challenging to create a less ambiguous diagram only based on the text description.
\end{remark}

\subsection{ Recurring simplex collaboration} 

The ``Recurring simplex collaboration'' is illustrated in Fig.~\ref{fig2}(b)---information is communicated multiple times in the direction from human experts to ChatGPT. In this case, the expert team begins with providing ``AEB System" and ``STPA Method" to ChatGPT, and then proceeds to perform each of the five steps of STPA. At each step, the experts pose the question to ChatGPT according to the steps of STPA.
Finally, ChatGPT outputs the overall result of the STPA analysis.

The first step of STPA involves ``defining accidents/hazards" (\textit{Q\&A 2.1}), where ChatGPT provides a list of accidents/hazards defined at the whole system level (i.e., cars in our AEB case study) due to \gls{AEB} failures.

\begin{tcolorbox}[enhanced,breakable,title = {AEB Example Q\&A 2.1: STPA Step 1}]
	\textbf{Q:} \textit{Could you help me define the accidents/hazards of cars due to AEB systems as the first step of STPA?}
	\tcblower
	\indent\textbf{A:} 
	(a) False Positives: AEB systems may mistake harmless obstacles or situations for potential collisions and trigger the brakes unnecessarily, causing the vehicle to stop suddenly and unexpectedly. This can result in rear-end collisions or accidents with other vehicles or pedestrians.
	
	(b) False Negatives: AEB systems may fail to detect potential collisions due to sensor malfunctions, environmental factors (e.g., adverse weather conditions), or human error (e.g., the driver's inattention or distraction). This can lead to accidents that could have been avoided if the system had functioned correctly.
	
	(c) Over-reliance on AEB: Drivers may become overly reliant on AEB systems, leading to a false sense of security and a decrease in their situational awareness. This can result in accidents when the AEB system is unable to prevent a collision or when the driver fails to react appropriately to a hazard.
	
	(d) Insufficient AEB Performance: AEB systems may not perform as intended in certain situations, such as high-speed or low-light conditions. This can lead to accidents if the system fails to brake in time or if the braking distance is longer than expected.
	
	(e) Malfunctioning AEB: AEB systems may malfunction due to software errors, hardware failures, or maintenance issues. This can cause the system to either trigger false positives or false negatives, or to fail altogether, leading to accidents.
\end{tcolorbox}

\begin{tcolorbox}[enhanced,breakable,title = {DSM Example Q\&A 2.1: STPA Step 1}]
	\textbf{Q:} \textit{Could you help me define the accidents/hazards of the Electricity Demand Side Management (DSM) system as the first step of STPA?}
	\tcblower
	\indent\textbf{A:} 
    The Systems-Theoretic Process Analysis (STPA) is a safety engineering methodology that focuses on identifying potential hazards and unsafe control actions within complex systems. The first step in STPA usually involves defining the system's purpose and the hazards or unsafe states that could lead to accidents. Once these are defined, the next steps look at control loops and unsafe control actions that could contribute to the system reaching one of these hazardous states.

    In the context of Electricity Demand Side Management (DSM), which aims to manage and optimize energy consumption on the consumer side, hazards could arise from various factors including but not limited to:

    Accidents/Hazards:
    (a) Power Surges or Spikes: 
    Unplanned changes in electricity demand can lead to power surges, affecting both the grid and consumer appliances.

    (b) System Overload:
    Improper management of energy demand could overwhelm the system, potentially causing failure in the grid.
    
    (c) Data Security Incidents:
    DSM often relies on real-time data. Unauthorized access could compromise the stability of the system.
    
    (d) Inaccurate Predictive Analysis:
    Inaccurate demand forecasting could result in inefficient load distribution.
    
    (e) Uncontrolled Shutdowns:
    Sudden, unplanned disconnection from the grid could lead to both minor inconveniences and major disruptions.
    
    (f) Rate Instabilities:
    Improper demand-side management could lead to fluctuating electricity rates, impacting consumers.
    
    (g) Resource Depletion:
    Poor DSM can lead to inefficient use of electricity, which can be especially problematic in areas where electricity generation relies on non-renewable resources.
    
    (h) Reduced Power Quality:
    If DSM systems cannot adapt to variations in energy production (such as from renewable sources), the quality of the delivered power may be compromised.
    
    (i) Economic Loss:
    Inefficient demand-side management can have economic consequences, such as increased operational costs.
    
    (j) Environment Hazards:
    Poorly managed systems may lead to increased emissions from electricity generation, affecting local and global environments.
    
    (k) Human Errors:
    Misconfiguration or lack of understanding of the DSM system by operators or consumers can lead to hazardous conditions.
    
    (l) Communication Failures:
    Failures in the communication system can lead to incorrect or delayed commands, causing system instability.
\end{tcolorbox}

Typically, traditional STPA focuses on identifying high-level consequences, such as traffic accidents or rear-end collisions in Baseline \Rmnum{1}, during its initial stage. 

\begin{remark}[Unrobustness to question phrased]
    If we phrase the question to ChatGPT in a slightly different way while keeping all the important keywords, the answer may also be different. If we substitute the question in \textit{AEB Example Q\&A 2.1} with ``What are the potential accidents and hazards that can occur in cars equipped with AEB systems, and how can we identify them as the first step of STPA?'' remaining the semantics and key words, we might receive different answers with a different number of hazards/accidents defined. 
\end{remark}

\begin{tcolorbox}[enhanced,breakable,title ={AEB Example Q\&A 2.2: STPA Step 2}]
	\textbf{Q:} \textit{Could you help me draw the model control loop structure in the AEB systems of cars as the second step of STPA?}
	\tcblower
	\textbf{A:} 
	(a) Sensors: The sensors in the AEB system include cameras, radar, lidar, and other sensors that detect potential collisions. These sensors send signals to the controller indicating the location and speed of nearby objects.
	
	(b) Controller: The controller in the AEB system receives the signals from the sensors and calculates the appropriate response to avoid or mitigate a collision. The controller determines whether to trigger the brakes, how much braking force to apply, and for how long.
	
	(c) Actuator: The actuator in the AEB system is responsible for physically applying the brakes. When the controller determines that braking is necessary, it sends a signal to the actuator, which applies the brakes to slow down or stop the vehicle.
	
	(d) Plant: The plant in the AEB system is the vehicle itself, including the braking system and other components that affect the behavior of the vehicle. The plant responds to the actions of the actuator by slowing down or stopping the vehicle.
\end{tcolorbox}

\begin{tcolorbox}[enhanced,breakable,title ={DSM Example Q\&A 2.2: STPA Step 2}]
	\textbf{Q:} \textit{Could you help me draw the model control loop structure in the Electricity Demand Side Management (DSM) system as the second step of STPA?}
	\tcblower
	\textbf{A:} 
	Entities in Control Loop:
        Controller: Energy Management System (EMS) / DSM Algorithm
        Controlled Process: Electrical Grid and Consumer Devices (HVAC, lights, etc.)
        Actuators: Smart Meters, Switches, and Relays for load shedding or on-off control
        Sensors: Energy Usage Monitors, Real-time Pricing Sensors
        Feedback: Energy Consumption Data, Real-time Pricing Data

    Control Commands
        The EMS/DSM Algorithm sends control commands to adjust energy consumption.
        Examples of Control Actions:
            Increase/Decrease HVAC temperature
            Turn off/on non-essential lights
            Load shedding during peak demand
    Actuation
        Actuators (smart meters, switches, and relays) execute the control commands on the consumer devices or grid elements.
    Controlled Process Changes
        The controlled processes (electrical grid and consumer devices) respond to the control commands, altering their behavior.
    Measurement and Monitoring
        Sensors collect data on the current state of the electrical grid and consumer devices. This could include power usage, current load, device states, etc.
    Feedback to Controller
        The sensors send this data back to the EMS/DSM Algorithm as feedback.
    Adjustment and Optimization
        The controller analyzes the feedback to adjust its control strategy and actions as required.

This model can be extended to include additional controllers and controlled processes, such as:

    Utility companies that are overseeing larger grid behavior
    Individual consumers or building management systems that have their own localized control strategies
    Government agencies that may have control over larger-scale electricity pricing or generation policies
    Each of these can form nested or parallel control loops interacting with each other. Identifying unsafe control actions within these loops would be the next step in the STPA process.
\end{tcolorbox}

% We arrange it into a figure based on ChatGPT's output, cf. Remark \ref{rm_no_graph}, as shown in Fig.~\ref{fig3}(b). 
% Upon comparison with Fig.~\ref{fig1}, it becomes evident that there exists a disparity between the output of ChatGPT and the block diagram of the \gls{AEB} system consolidated by the expert team. To be more specific, in \textit{Case 2}, the ChatGPT output from \textit{Example Q\&A 2.2} may be incomplete, lacking certain components due to without expert knowledge adjustment. For instance, the output contains no information regarding drivers, environmental information, and other related components. To compare with the baseline \cite{10.1115/1.4051940} (which prescribes the ``implementation of brake'' as the example), without analysing all components in the answer of \textit{Example Q\&A 2.2}, we only choose the option (c) as the example module for continuing the safety analysis in the subsequent steps.
In the \textit{Recurring simplex collaboration} case, the ChatGPT output from \textit{AEB Example Q\&A 2.2} may be incomplete, lacking certain components due to without expert knowledge adjustment. For instance, the output contains no information regarding drivers, environmental information, and other related components. To compare with the baseline \cite{10.1115/1.4051940} (which prescribes the ``implementation of brake'' as the example), without analysing all components in the answer of \textit{AEB Example Q\&A 2.2}, we only choose the option (c) as the example module for continuing the safety analysis in the subsequent steps.

\begin{tcolorbox}[enhanced,breakable,title ={AEB Example Q\&A 2.3: STPA Step 3 (on the braking module)}]
	\textbf{Q:} \textit{Could you help me dentify Unsafe Control Actions based on braking module in the AEB systems of cars as the third step of STPA?}
	\tcblower
	\textbf{A:} 
	(a) Delayed Braking: If the braking module does not apply the brakes quickly enough in response to a potential collision, the vehicle may collide with the obstacle or pedestrian. This could occur due to a malfunctioning sensor, a slow response from the controller, or a slow actuation of the brakes.
	
	(b) Over-application of Brakes: If the braking module applies too much braking force, the vehicle may come to an abrupt stop, leading to rear-end collisions or loss of control. This could occur due to incorrect calibration of the braking system or a malfunctioning actuator.
	
	(c) Under-application of Brakes: If the braking module does not apply enough braking force, the vehicle may not slow down enough to avoid a collision, leading to accidents. This could occur due to faulty sensor data or incorrect calculations by the controller.
	
	(d) False Braking: If the braking module applies the brakes unnecessarily, such as in response to a false positive from the sensors, the vehicle may come to an abrupt stop or swerve unexpectedly, leading to accidents or collisions. This could occur due to incorrect calibration of the sensors or a malfunctioning controller.
	
	(e) Inadequate Braking: If the braking module does not provide enough braking power, such as in high-speed or heavy-load conditions, the vehicle may not slow down enough to avoid a collision, leading to accidents or collisions. This could occur due to inadequate design or specifications of the braking system.
\end{tcolorbox}

\begin{tcolorbox}[enhanced,breakable,title ={DSM Example Q\&A 2.3: STPA Step 3 (Control action: Excess capacity demand)}]
	\textbf{Q:} \textit{Could you help me dentify Unsafe Control Actions based on the control action (Excess capacity demand) in the DSM systems as the third step of STPA?}
	\tcblower
	\textbf{A:} 
    (a) Providing No Response to Excess Capacity Demand
        Failure to act when there is an excess capacity demand could lead to grid instability, blackouts, or brownouts.

    (b) Excessive Load Shedding
        Overreacting to excess demand by shedding too much load could create disruptions and affect critical infrastructure like hospitals, data centers, etc.

    (c) Incorrect Timing
        Responding too early or too late to excess capacity demand can either unnecessarily restrict usage or allow dangerous spikes.

    (d) Partial or Incomplete Execution
        Not fully executing a load reduction when needed can still result in an overloaded system.

    (e) Miscalculating the Demand
        Incorrect assessment of excess capacity demand can lead to inappropriate control actions, such as excessive or insufficient load shedding.

    (f) Inaccurate Targeting
        Incorrectly identifying which consumers or devices should reduce load can result in unnecessary inconvenience or even hazardous conditions.

    (g) Command Conflicts
        Conflicting control actions (e.g., a local controller attempting to increase load while the central controller tries to shed load) can create an unstable system.

    (h) Ignoring Priority Levels
        Not considering the priority levels of different consumers (e.g., critical infrastructure) during load shedding can lead to safety risks.

    (i) Improper Feedback Interpretation
        Misinterpreting sensor feedback can lead to inappropriate control actions, exacerbating the problem instead of solving it.

    (j) Failure to Communicate
        Not communicating the control action effectively to the actuators or end-users could result in non-compliance or unexpected behavior.
    (k) Rate Instability
        Reacting to excess capacity demand by fluctuating electricity rates without proper planning can lead to economic instability.
    (l) Security Risks
        Unauthorized manipulation of control actions could occur, leading to a compromised and unsafe state.
\end{tcolorbox}

% There are 5 \gls{UCAs} of the braking module in the AEB system identified in \textit{Example Q\&A 2.3}. Specifically, answer (a) associates to ``UCA-3'' in Table \ref{tab1}, while answers (b) and (c) associate to ``UCA-4'', answer (d) associates to ``UCA-2'', and answer (e) appears to be similar to answer (c). 
% % By employing the recurring simplex interaction in \textit{Case 2}, 
% It is observed that the outcomes presented in \textit{Example Q\&A 2.3} already encompass ``UCA-2'', ``UCA-3'', and ``UCA-4'' of the baseline. Thus, they convey the same safety implication at the safety semantic level, although there may be differences in their presentation and methodology, cf. Remark \ref{rm_unreli}.
It is observed that the outcomes presented in \textit{AEB Example Q\&A 2.3} already encompass ``The brake command is inadequately provided under no collision
risk'', ``Brake command occurring too late causes insufficient brake and crash risk'', and ``Braking too long or too short causes a hazard'' of the baseline. Thus, they convey the same safety implication at the safety semantic level, although there may be differences in their presentation and methodology, cf. Remark \ref{rm_unreli}.

\begin{remark}[Precise answers from specific questions]
    Upon comparing the \gls{UCAs} outlined in \textit{AEB Example Q\&A 1.2} and \textit{AEB Example Q\&A 2.3}, it can be observed that when the question is more specific, e.g., ``identify unsafe control actions based on braking module'', the analysis results are more focused and precise. 
\end{remark}

In the Baseline \Rmnum{2} case study, the output for \textit{DSM Example Q\&A 2.1} essentially covers the accidents or losses that could be caused by the DSM system, as outlined in Table \ref{tab: accidents}, along with potential hazards. However, the response to \textit{DSM Example Q\&A 2.2} merely contains the basic components of the elementary DSM system, basically aligning with the control loop structure depicted in Figure \ref{fig:middle_loop}.
Upon further inquiry into the UCAs associated with control actions, we find that subsections (a), (b), (c), (d), and (e) in \textit{DSM Example Q\&A 2.3} generally align with the conclusions reached by the expert group, as displayed in Table \ref{tab: accidents}. ChatGPT even identifies potential UCAs that the expert team ought to consider, such as in \textit{DSM Example Q\&A 2.3} (g), (k), and (l). Additionally, it accounts for other UCAs as illustrated in \textit{DSM Example Q\&A 2.3} (f), (i), and (j).

Given that the baselines do not emphasise identifying causal scenarios and safety requirements in the paper \cite{10.1115/1.4051940,karatzassystem}, we have chosen to omit the outcomes of these two stages in our discussion.

\subsection{Recurring duplex collaboration}

As the name suggests, the third case integrating ChatGPT in STPA involves multiple bi-directional interactions. As shown in Fig.~\ref{fig2}(c), this \textit{Recurring duplex collaboration} initiated by the human safety experts, sending ``AEB System'' and ``STPA Method" to ChatGPT.
However, unlike \textit{Recurring simplex collaboration} in Fig.~\ref{fig2}(b), they collect each output of ChatGPT, analyse the information by incorporating domain knowledge, and then query ChatGPT in turn based on the human processed question. The expert team is responsible for ``Sending Messages", ``Reserving Messages", and ``Analysing Messages" in each step of STPA.

As the \textit{Recurring duplex collaboration} case involves multiple bi-directional collaborations, the expert team is required to analyse the output of ChatGPT with expert knowledge after each step of STPA. In the following \textit{Example Q\&A}, \textbf{A1} denotes the response given by ChatGPT, while \textbf{A2} is the outcome of the expert team after analysing \textbf{A1} with their domain knowledge. 

\begin{tcolorbox}[enhanced,breakable,title = {AEB Example Q\&A 3.1: STPA Step 1}]
	\textbf{Q:} \textit{Could you help me define the accidents/hazards of cars due to AEB systems as the first step of STPA?}
	\tcblower
	\indent\textbf{A1:}
	(a) False activations: AEB systems can sometimes mistakenly detect an object in the car's path and activate the brakes, even when there is no danger of a collision. This can result in sudden, unexpected stops that could cause accidents, especially if the driver is following too closely.
	(b) Missed activations: On the other hand, AEB systems may not activate when they should, such as when the system fails to detect an obstacle in the car's path. This can result in a collision that the AEB system was supposed to prevent.
	(c) Inappropriate speed reduction: AEB systems may cause the car to slow down too abruptly or to an inappropriate speed, leading to a collision with a vehicle following behind or causing other drivers to swerve to avoid a collision.
	(d) Malfunction: AEB systems are complex systems that rely on sensors, cameras, and other components to function correctly. If any of these components malfunction, the AEB system may fail to detect obstacles or activate the brakes in time.
	(e) Driver over-reliance: If drivers become too reliant on AEB systems, they may become complacent or fail to pay attention to the road, which could lead to accidents.\\

\end{tcolorbox}

\begin{tcolorbox}[enhanced,breakable]
	\textbf{A2 (by safety experts after considering the ChatGPT answer of AEB Q\&A 3.1):} 
	\tcblower
	(a) Rear-end collisions: AEB systems may engage the brakes unexpectedly, such as when detecting a false positive or when the driver is still in control of the vehicle. This could lead to a rear-end collision with another vehicle, especially if the driver behind the vehicle with AEB is not prepared for sudden braking.
	(b) Pedestrian or cyclist accidents: AEB systems may not always detect pedestrians or cyclists, particularly in low-light conditions or when the object is not moving. This can lead to potential accidents if the driver is relying on the AEB system to detect all hazards and does not notice the pedestrian or cyclist.
	(c) Delayed response: AEB systems may have a delay in detecting obstacles, causing the system to initiate braking too late, resulting in a collision.
	(d) Software bugs: AEB systems may contain software bugs that cause the system to malfunction, leading to incorrect or delayed responses.
	(e) Malfunction due to environmental factors: AEB systems may malfunction due to environmental factors, such as rain, fog, or snow, that affect the sensors' ability to detect obstacles.
	(f) False activations: AEB systems can sometimes mistakenly detect an object in the car's path and activate the brakes, even when there is no danger of a collision.
\end{tcolorbox}

\begin{tcolorbox}[enhanced,breakable,title = {DSM Example Q\&A 3.1: STPA Step 1}]
	\textbf{Q:} \textit{Could you help me define the accidents/hazards of the Electricity Demand Side Management (DSM) system as the first step of STPA?}
	\tcblower
	\indent\textbf{A1:}
Accidents:
    Power Outages: Inadequate demand management could lead to a situation where the grid cannot handle the load, causing outages.
    Equipment Failure: Incorrect signaling or incentives may cause electrical devices to malfunction.
    Economic Loss: Poorly implemented DSM could lead to economic inefficiencies, raising costs for utilities or consumers.
    Data Breaches: Unauthorized access to the system data.
    Health and Safety Risks: Any situation where the electricity supply is so mismanaged that it leads to accidents like fires or injuries.
Hazards:
    Inaccurate Forecasting: Failing to accurately predict electricity demand.
    System Overload: Exceeding the capacity of the grid.
    Improper Communication: Miscommunication between the utility and the end-users.
    Timing Errors: Incorrect timing of lowering or increasing energy demands.
    Rate Inconsistencies: Confusing or inconsistent pricing strategies that don't accurately reflect demand.
    Human Errors: Errors in system monitoring, either by the consumers or the utility.
    Unauthorized Access: Risk of hacking or unauthorized control of the DSM system.
    Data Integrity: Corruption or loss of critical system data.
    Dependency Risks: Risks arising from interdependencies with other systems like weather forecasting, grid management, etc.
    Regulatory Risks: Risks stemming from not meeting regulatory or legal guidelines on energy management, data protection, etc.
\end{tcolorbox}

\begin{tcolorbox}[enhanced,breakable]
	\textbf{A2 (by safety experts after considering the ChatGPT answer of DSM Q\&A 3.1):} 
	\tcblower
    (a) Power Outages: Inadequate demand management could lead to a situation where the grid cannot handle the load, causing outages.
    (b) Equipment Failure: Incorrect signaling or incentives may cause electrical devices to malfunction.
    (c) Economic Loss: Poorly implemented DSM could lead to economic inefficiencies, raising costs for utilities or consumers.
    (d) Data Breaches: Unauthorized access to the system data.
    (e)Health and Safety Risks: Any situation where the electricity supply is so mismanaged that it leads to accidents like fires or injuries.
\end{tcolorbox}

\begin{remark}[Irreproducibility]
    ChatGPT's responses may vary even if the exact same question (in the same context) is asked repeatedly (e.g. \textit{AEB Example Q\&A 2.1} and \textit{AEB Example Q\&A 3.1 (A1)})\footnote{Although it is known that ChatGPT provide answers to questions considering the context, the variation of answers to \textit{AEB Examples Q\&A 2.1} and \textit{AEB Q\&A 3.1} is not due to different contexts given both are asked at the first step of STPA. This is unsurprising given GPT produces a probability distribution over the set of possible next words or characters, thus identical inputs may not necessarily yield the same outputs.}. This is because ChatGPT utilises machine learning algorithms that generate responses based on probabilistic predictions.
\end{remark}

\begin{tcolorbox}[enhanced,breakable,title = {AEB Example Q\&A 3.2: STPA Step 2}]
	\textbf{Q:} \textit{Could you help me draw the model control loop structure in the AEB systems of cars as the second step of STPA?}
	\tcblower
	\textbf{A1:} 
	(a) Inputs: The AEB system receives input from various sensors, such as radar, lidar, and cameras, to detect obstacles in the car's path.
	(b) Perception: The AEB system processes the input data to perceive the environment and identify potential obstacles.
	(c) Decision-making: Based on the perceived environment, the AEB system makes a decision on whether to activate the brakes to avoid a collision.
	(d) Outputs: If the AEB system decides to activate the brakes, it sends a signal to the car's brake system to apply the brakes.
	(e) Effects: The AEB system's output (brake activation) affects the car's motion, potentially avoiding a collision or causing an unexpected stop.
	(f) Feedback: The AEB system may receive feedback from the car's sensors or other components, such as the brake system or steering system, to adjust its perception or decision-making process.
	(g) Human interaction: The driver may interact with the AEB system by overriding its decisions or disabling the system altogether.\\
	
\end{tcolorbox}

\begin{tcolorbox}[enhanced,breakable]
	\textbf{A2 (by safety experts after considering the ChatGPT answer of AEB Q\&A 3.2):}
	\tcblower
	(a) Driver: They play a critical role in determining whether or not the system is activated.
	(b) HMI: HMI stands for Human Machine Interface, which refers to the interface through which a human operator interacts with a machine or system
	(c) Environmental information: It allows the system to detect potential hazards and respond accordingly. Environmental information can also help the AEB system to adjust its response based on the driving conditions.
	(d) Senseors: The AEB system receives input from various sensors, such as radar, lidar, and cameras, to detect obstacles in the car's path.
	(e) Object tracker: The AEB system processes the input data to perceive the environment and identify potential obstacles.
	(f)Control unit: Based on the perceived environment, the AEB system makes a decision on whether to activate the brakes to avoid a collision.
	(g) Brakes: If the AEB system decides to activate the brakes, it sends a signal to the car's brake system to apply the brakes.
	(h) Power: The power support is a critical component of the AEB system as it ensures that the brakes are applied with sufficient force to avoid or mitigate a collision.
	(i) Feedback: The AEB system may receive feedback from the car's sensors or other components, such as the brake system or steering system, to adjust its perception or decision-making process.
\end{tcolorbox}

\begin{tcolorbox}[enhanced,breakable,title = {DSM Example Q\&A 3.2: STPA Step 2}]
	\textbf{Q:} \textit{Could you help me draw the model control loop structure in the Electricity Demand Side Management (DSM) system as the second step of STPA?}
	\tcblower
	\textbf{A1:} 
Components of Control Loop:
    Controller:

    This is typically the DSM system software that makes decisions based on various inputs.
    Controlled Process:

    The electrical grid along with consumer electrical devices.
    Actuators:

    These could be automated switches, load balancers, or programmable devices that can adjust electrical consumption.
    Sensors:

    Smart meters, grid monitors, and other devices that measure electricity consumption, grid health, etc.
    Feedback:
    
    The data that flows back into the controller to inform it about the state of the controlled process.

Control Loop Structure:
        Set Point:
        
        Desired level of electricity consumption, grid health, etc., as determined by policy or real-time analysis.
        Process Model:
        
        The internal model in the controller that predicts outcomes based on various actions.
        Decision Algorithms:
        
        Algorithms that decide what actions should be taken (e.g., load-shifting, turning off devices, altering rates).
        Control Commands:
        
        Instructions sent from the Controller to Actuators to modify the state of the Controlled Process.
        Process Data:
        
        Data from Sensors that describe the current state of the Controlled Process.
        Feedback Loop:
        
        Pathway through which Process Data is sent back to the Controller to update the Process Model and make new decisions.
\end{tcolorbox}

\begin{tcolorbox}[enhanced,breakable]
	\textbf{A2 (by safety experts after considering the ChatGPT answer of DSM Q\&A 3.2):}
	\tcblower
	The D/R Manager sends a Capacity request to the Distribution Network Operator (DNO), and the DNO returns the Capacity to the D/R Manager. D/R Manager continues to send Capactity Limit to Load Balancer and then the feedback is Capacity. Load Balancer sends Scheduled Load to Load Forecaster and then receives feedback that it is Predicted Load. Load Forecaster sends feedback information to D/R Manager. Load Balancer also sends Available Capacity to Admission Controller and then Admission Controller sends Reject Request to Load Balancer. Universal App. Controller sends Load Request to Admission Controller, and then Admission Controller returns Accept Request. Load Balancer also sends Scheduled Load to Universal App. Controller. Universal App. Controller also sends Operation Commands like Adaptors, and then Adaptors sends Actuate Commands (Start, stop, etc.) to Smart Appliances, and then Smart Appliances sends Elastic Loads to Smart Meters, and Smart Meters sends Consumption Metering back to Universal App. Controller . Smart Applicances also send Inelastic Loads and Operational State to the App. Interface, and then the App. Interface returns the Operational Status to the Universal App. Controller. App.Interface also returns Preemption, required energy, heuristic value and nominal power to Load Balancer. App. Interface also sends feedback to Comfort Context. Smart Appliances also send Settings values to Comfort Context. Env. Sensors obtains Conditions from Environment and then sends control actions to Comfort Context. Comfort Context will return Comfort Boundaries to Comfort Context. Please apply Systems-Theoretic Process Analysis (STPA) to this system and provide complete STPA analysis results.

\end{tcolorbox}

% The incorporation of specialised expertise is a vital aspect of constructing the model of control loop structure, as the diagram illustrated in Fig. \ref{fig3}(c) based on \textit{Case 3}. 
% \textit{Example Q\&A 3.2 (A1)} shows that, while the ChatGPT is able to generate the control loop structure, there is the need for a more comprehensive summary and refinement by the human expert team.

\begin{tcolorbox}[enhanced,breakable,title = {AEB Example Q\&A 3.3: STPA Step 3} (on the braking module)]
	\textbf{Q:} \textit{Could you help me dentify unsafe control actions based on braking module in the AEB systems of cars as the third step of STPA?}
	\tcblower
	\textbf{A1:} 
	(a) Delayed braking: If the AEB system detects an obstacle but there is a delay in the activation of the braking module, the car may not stop in time to avoid a collision.
	(b) Over-reliance on braking: If the AEB system relies solely on the braking module to avoid a collision, it may not take into account other factors that could affect the car's trajectory, such as the road conditions or the car's speed.
	(c) Inappropriate braking force: If the AEB system applies too much or too little force to the brakes, it could cause the car to skid or lose control.
	(d) Failure to disengage: If the AEB system fails to disengage the braking module after a collision is avoided, it could cause the car to come to an unexpected stop, potentially causing a hazard to other drivers.
	(e) Incorrect target identification: If the AEB system incorrectly identifies the target object, it may apply the brakes unnecessarily, leading to unexpected stops or collisions.\\
\end{tcolorbox}

\begin{tcolorbox}[enhanced,breakable]
	\textbf{A2 (by safety experts after considering the ChatGPT answer of AEB Q\&A 3.3):}
	\tcblower
	(a) Braking Failure: If the braking module malfunctions and fails to apply the brakes at all, it could lead to a collision and potential injury or damage.
	(b) Abrupt Braking: If the AEB system applies the brakes abruptly, it could cause the vehicle to skid or lose control, leading to a potential accident.
	(c) Overly Sensitive Braking: If the AEB system is too sensitive and initiates braking too frequently, it could cause unnecessary wear and tear on the brakes and reduce the vehicle's performance.
	(d) Incorrect classification of the obstacle: If the AEB system incorrectly identifies the target object, it may apply the brakes unnecessarily, leading to unexpected stops or collisions.
	(e) Delayed braking: If the AEB system detects an obstacle but there is a delay in the activation of the braking module, the car may not stop in time to avoid a collision.
	(f) Inappropriate braking force: If the AEB system applies too much or too little force to the brakes, it could cause the car to skid or lose control.
\end{tcolorbox}

\begin{tcolorbox}[enhanced,breakable,title = {DSM Example Q\&A 3.3: STPA Step 3} (Control action: Excess capacity demand)]
	\textbf{Q:} \textit{Could you help me dentify Unsafe Control Actions based on the control action (Excess capacity demand) in the DSM systems as the third step of STPA?}
	\tcblower
	\textbf{A1:} 
Not Providing Excess Capacity Demand When Needed
If the DSM system fails to signal or implement an excess capacity demand when it is actually needed (e.g., during peak times to prevent blackouts), that could be hazardous.

Providing Excess Capacity Demand When Not Needed
If the DSM system signals excess capacity demand when it isn't needed, it could lead to inefficiencies, higher costs, and may also stress the grid unnecessarily.

Providing Excess Capacity Demand Too Late
If the DSM system reacts too slowly in flagging an excess capacity demand, the grid might already be in a stressed state, and the delayed action could be ineffective or even counterproductive.

Providing Excess Capacity Demand Too Early
If the action is taken too early, it may result in unnecessary adjustments that can lead to inefficiencies and increased operational costs.

Inconsistent Excess Capacity Demand Signals
If the DSM system sends inconsistent signals, such as rapidly toggling between different states, it could confuse the actuators or human operators, leading to ineffective demand management.

Excessive Amounts of Excess Capacity Demand
Overcompensating by requiring an excessive amount of additional capacity could lead to operational inefficiencies and may overload certain components of the system, leading to premature wear and tear.

Insufficient Amounts of Excess Capacity Demand
Underestimating the additional capacity needed could still lead to grid stress, blackouts, or other safety issues.

Unauthenticated or Unauthorized Excess Capacity Demand
If the control action is not properly authenticated, it opens the door for unauthorized or malicious entities to manipulate the system.

Lack of Communication or Confirmation for Excess Capacity Demand
If the control action isn't confirmed by actuators or secondary controllers, the DSM system might assume the action has been taken when it hasn't, leading to potential risks.

Incorrect Geographic or Sectoral Application of Excess Capacity Demand
If the DSM system applies the control action to the wrong geographic area or consumer sector (industrial vs. residential), the benefits might not be realized, and it could exacerbate existing problems.
\end{tcolorbox}

\begin{tcolorbox}[enhanced,breakable]
	\textbf{A2 (by safety experts after considering the ChatGPT answer of DSM Q\&A 3.3):}
	\tcblower
	(a) Providing Excess Capacity Demand When Unnecessary.
    (b) Not Providing Excess Capacity Demand When Necessary.
    (c) Providing Excess Capacity Demand Too Late.
    (d) Providing Excess Capacity Demand Too Early.
    (e) Providing Excess Capacity Demand to the Wrong Locations.
    (f) Providing Excess Capacity Demand Without Proper Authorization or Security Checks.
    (g) Providing Excess Capacity Demand in Conflict with User Requirements.
    (h) DRM does not provide capacity limits when these have been modified.
    (i) Failure to Coordinate with Other Systems.
    (j) Failure to Override.
\end{tcolorbox}

In the context of \textit{AEB Example Q\&A 3.3}, ``No brake command provides a cause of a hazard'' in Table \ref{tab: accidents_aeb} corresponds to A2(d), while A1(e) and A2(b) may both refer to ``The brake command is inadequately provided under no collision
risk''. A1(a) conveys the safety semantics of ``Brake command occurring too late causes insufficient brake and crash risk'', and A1(c) corresponds to ``Braking too long or too short causes a hazard''. Notably, all the 4 UCAs in Table \ref{tab: accidents_aeb} are now covered by \textbf{A2}, thanks to the interactive collaboration of ChatGPT and human safety experts.

\begin{remark}[Comprehensibility from interactivity]
\label{rm_com_inter}
    By applying the recurring duplex collaboration case, it becomes evident that the four UCAs in the baseline are all identified. In addition, new UCAs are also discovered.
    %by the expert team can also be accommodated, as compared to the baseline example. 
    Increased interaction frequency between the expert team and ChatGPT allows the model to acquire additional domain knowledge, resulting in more detailed and informative answers.
\end{remark}

At the conclusion of each STPA step, the expert team will consolidate the answers. For instance, in this step, the expert team can determine ``Brake duration is either too long or too short.'' They can then use this information as input for the next step of STPA.
The baseline \Rmnum{2} case study further confirms that the A1 answer in \textit{DSM Example Q\&A 3.3} largely aligns with expert analysis, although it may not be exhaustive (cf. Remark \ref{rm_com_inter}). By integrating expert knowledge in A2 with ChatGPT's output, the final results become more clear and precise. 
The ChatGPT's understanding in the A1 of \textit{DSM Example Q\&A 3.1} almost covers the analysis results of the expert team, but it still cannot be completely and accurately described with the A1 of\textit{DSM Example Q\&A 3.2} system control loop structure (cf. Remark \ref{rm_no_graph}). The expert team can analyse and merge the analysis results and continue to query ChatGPT for the next step of STPA. In a similar vein, there is chosen to omit the discourse pertaining to the fourth and fifth stages of STPA.

\end{document}